\pdfoutput=1

\documentclass[11pt]{article}

\usepackage{EMNLP2023}

\usepackage{times}
\usepackage{latexsym}

\usepackage[T1]{fontenc}

\usepackage[utf8]{inputenc}
\usepackage{inconsolata}

\usepackage{microtype}

\usepackage{enumitem}
\usepackage{url}
\usepackage{arydshln}
\usepackage{graphicx}
\usepackage{booktabs}
\usepackage{pifont}

\usepackage{algorithm}
\usepackage[noend]{algpseudocode}
\usepackage{xcolor, soul}
\usepackage{multirow, makecell}

\definecolor{Background}{RGB}{217,245,203}
\sethlcolor{Background}

\definecolor{Back_Purple}{RGB}{238,215,236}
\definecolor{Back_Yellow}{RGB}{250,239,197}

\newif\ifcomments
\commentstrue 
\ifcomments
    \newcommand\oy[1]{\textcolor{red}{[OY: #1]}}
    \newcommand\tw[1]{\textcolor{magenta}{[TW: #1]}}
    \newcommand\jb[1]{\textcolor{blue}{[JB: #1]}}
    \newcommand\dd[1]{\textcolor{magenta}{[DD: #1]}}
    \newcommand\bb[1]{\textcolor{brown}{[BB: #1]}}
    \newcommand\uk[1]{\textcolor{green}{[UK: #1]}}

\else
    \providecommand{\oy}[1]{}
    \providecommand{\tw}[1]{}
    \providecommand{\bb}[1]{}
    \providecommand{\uk}[1]{}
    \providecommand{\jb}[1]{}
    \providecommand{\dd}[1]{}
\fi

\newcommand\bamboogle{\textsc{Bamboogle}}
\newcommand\quartz{\textsc{Quartz}}
\newcommand\hotpotqa{\textsc{HotpotQA}}
\newcommand\strategyqa{\textsc{StrategyQA}}

\newcommand\feverous{\textsc{Feverous}}
\newcommand\wikihop{\textsc{2WikiMQA}}
\newcommand\fermi{\textsc{Fermi}}

\newcommand\numdatasets{7}

\newcommand\selfask{$\textsc{SA}$}
\newcommand\selfconsistency{$\textsc{SC}$}
\newcommand\ste{$\textsc{SCR}$}
\newcommand\stec{$\textsc{SCR-Ev}$}
\newcommand\mte{$\textsc{MCR}$}
\newcommand\mtec{$\textsc{MCR-Ev}$}

 \newcommand{\vicuna}{\texttt{Vicuna-13B}}
  \newcommand{\codex}{\texttt{code-davinci-002}}
  \newcommand{\davinci}{\texttt{text-davinci-002}}

\newcommand{\retriever}{retriever}
\newcommand{\entailment}
{meta-reasoner}
\newcommand{\decomposition}{decomposition}

\newcommand{\fone}{F$_1$}

\title{Answering Questions by Meta-Reasoning over Multiple Chains of Thought}

\author{\makecell{Ori Yoran$^{*1}$~~~~~ Tomer Wolfson$^{*1,2}$~~~~~ Ben Bogin$^{1}$ ~~~~~ Uri Katz$^{3}$ \\~~~~~ Daniel Deutch$^{1}$ ~~~~~ Jonathan Berant$^{1}$ } \\ 
$^{1}$Tel Aviv University\hspace{5mm}
$^{2}$Allen Institute for AI \hspace{5mm} $^{3}$Bar Ilan University \\ 
\texttt{\small\makecell{ori.yoran@cs.tau.ac.il}} \texttt{\small\makecell{tomerwol@mail.tau.ac.il}}}

\begin{document}
\maketitle

\def\thefootnote{*}\footnotetext{Both authors contributed equally to this work.}\def\thefootnote{\arabic{footnote}}

\begin{abstract}

Modern systems for multi-hop question answering (QA) typically break questions into a sequence of reasoning steps, termed \emph{chain-of-thought} (CoT), before arriving at a final answer. Often, multiple chains are sampled and aggregated through a voting mechanism over the final answers, but the intermediate steps themselves are discarded. While such approaches improve performance, they do not consider the relations between intermediate steps across chains and do not provide a unified explanation for the predicted answer.
We introduce Multi-Chain Reasoning (\mte{}), an approach which prompts large language models to \emph{meta-reason} over multiple chains of thought, rather than aggregate their answers. 
MCR examines different reasoning chains, mixes information between them and selects the most relevant facts in generating an explanation and predicting the answer. 
\mte{} outperforms strong baselines on \numdatasets{} multi-hop QA datasets.
Moreover, our analysis reveals that \mte{} explanations exhibit high quality, enabling humans to verify its answers.

\end{abstract}

\section{Introduction}
\label{sec:introduction}

\begin{figure}[t]\setlength{\belowcaptionskip}{-8pt}
  \centering
  \includegraphics[trim={0cm 19.5cm 7.4cm 0cm}, clip, width=0.49\textwidth]{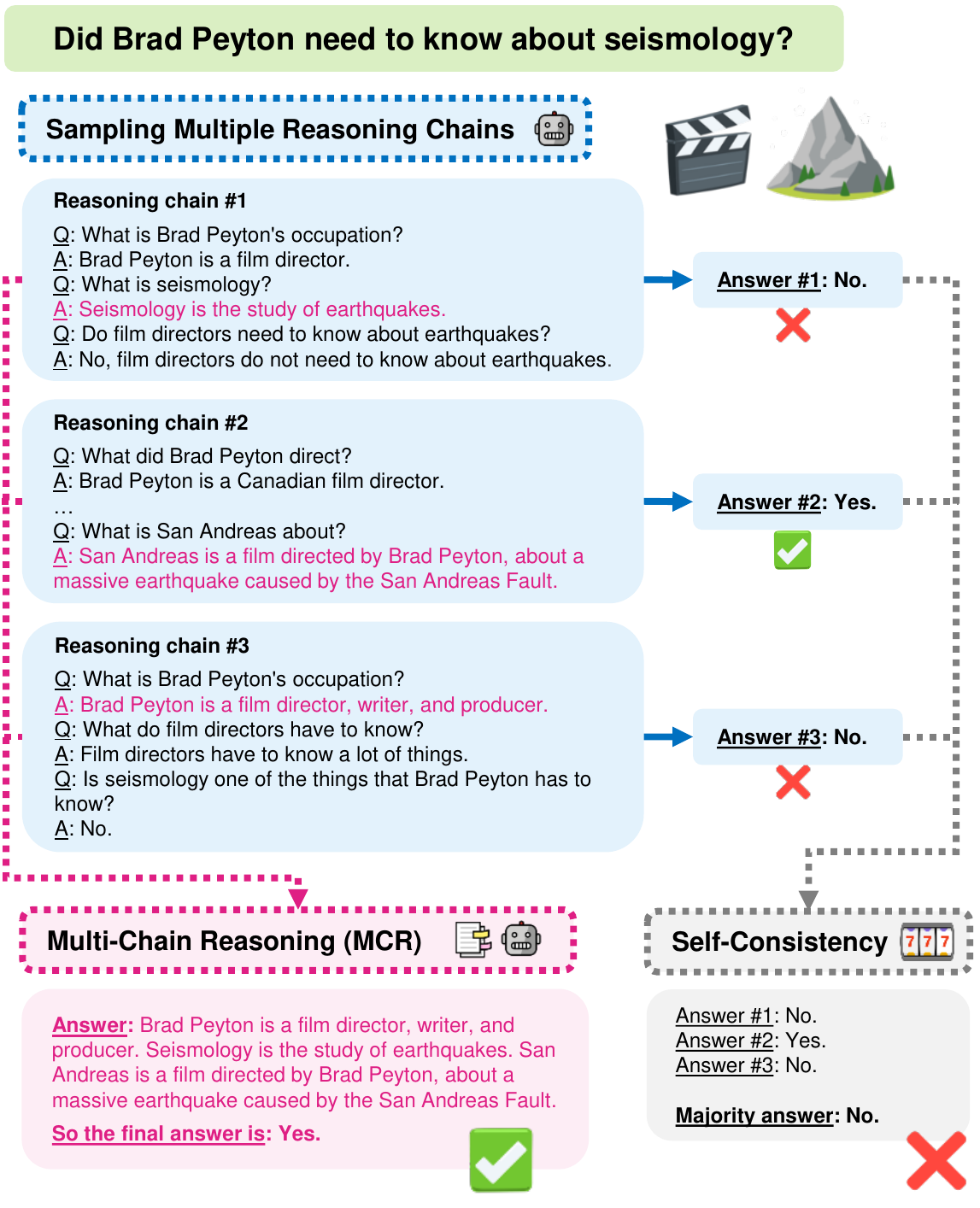}
  \vspace*{-0.7cm}
  \caption{An example from \strategyqa{}, showing the output of Multi-Chain Reasoning versus Self-Consistency. \mte{} uses reasoning chains as its \emph{context} for QA. \selfconsistency{} solely relies on the chains' answers.}
  \label{fig:multi_chain_example}
\end{figure}

\begin{figure*}[t]\setlength{\belowcaptionskip}{-8pt}
  \centering
  \includegraphics[trim={0cm 0cm 0.5cm 0cm}, clip, width=\textwidth]{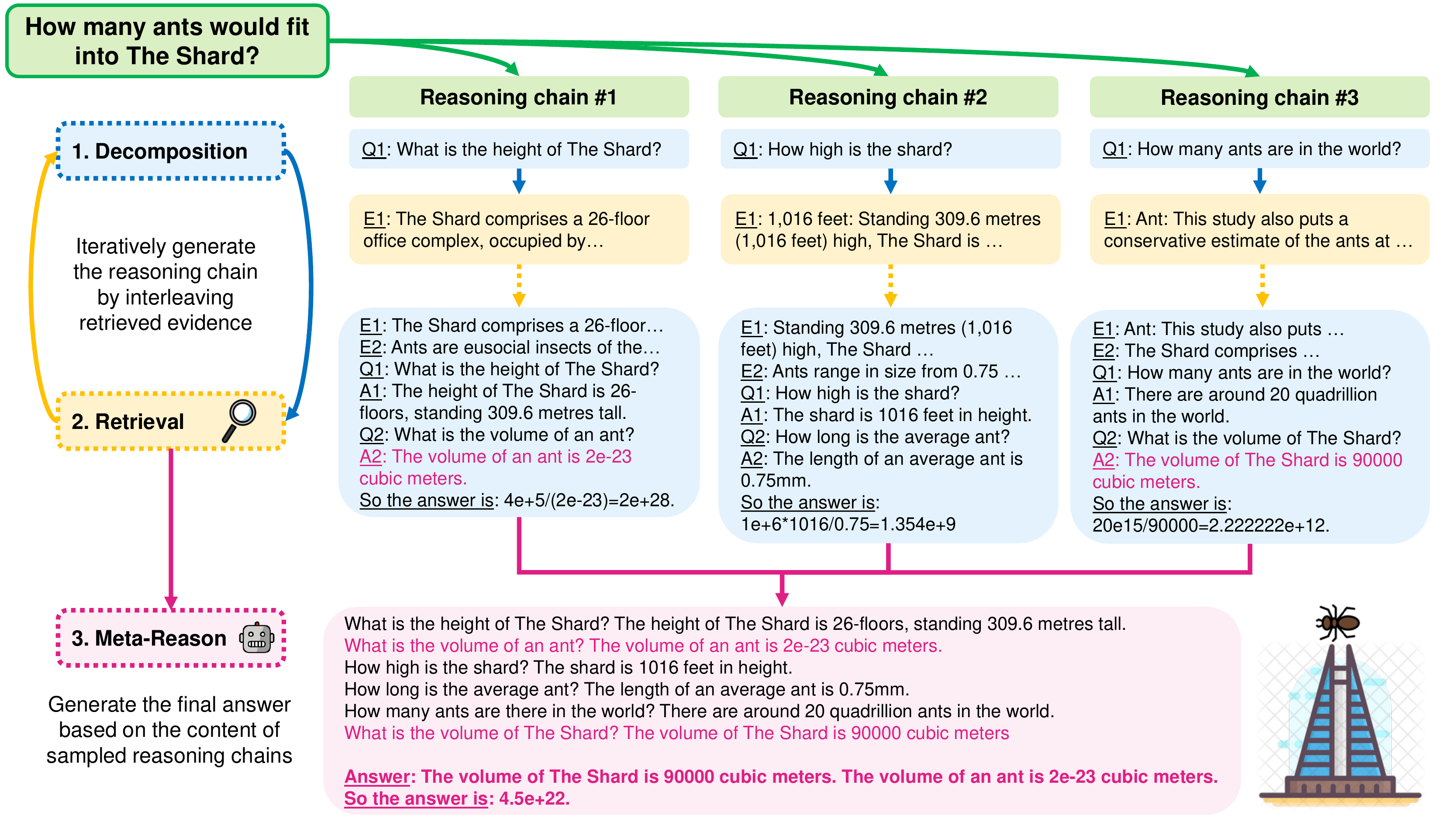}
  \vspace*{-0.7cm}
  \caption{An overview of \mte{},  given a question from the \fermi{} dataset. Steps 1-2 generate multiple reasoning chains by conditioning the generation of intermediate questions and answers on retrieved evidence sentences. In step 3, the \emph{\entailment{}} generates the final answer, given multiple reasoning chains from the previous steps.}
  \label{fig:architecture}
\end{figure*}

In chain-of-thought (CoT) prompting, a large language model \cite{NEURIPS2020_1457c0d6, Chowdhery2022PaLMSL,Kadavath2022LanguageM, Touvron2023LLaMAOA} is prompted to 
generate its answer following a step-by-step explanation \cite{Wei2022ChainOT, nye2022show}. CoT prompting has been shown to dramatically improve performance on reasoning-heavy tasks \cite{kojima2022large, Zhou2022LeasttoMostPE}. 
Furthermore, \citet{wang2023selfconsistency} showed that sampling \emph{multiple} chains of thought and returning their majority output further improves accuracy, a method which they term \emph{self-consistency} (SC).

While SC leads to performance gains, it also has several shortcomings. First, when the space of possible outputs is large \cite{kalyan-etal-2021-much}, each reasoning chain may lead to a different output, in which case no significant majority will be formed. Second, focusing exclusively on the final output discards relevant information that is present in the intermediate reasoning steps. 
Consider answering the question 
\emph{``Did Brad Peyton need to know about seismology?''} (Fig.~\ref{fig:multi_chain_example}). Reasoning chain \#1 leads to an incorrect answer (\textit{``No''}), but its steps provide useful information. For example, the intermediate question, and following answer, on \textit{``What is seismology?''} constitute an important fact that is absent from the other two chains.
Last, using SC jointly with chain-of-thought prompting reduces interpretability, as there is no single reasoning chain that can be considered as an explanation. 



In this work, we propose \emph{Multi-Chain Reasoning} (MCR), where we prompt a large language model (LLM) to \emph{meta-reason} across multiple reasoning chains and produce a final answer, alongside an explanation. Unlike prior work, sampled reasoning chains are used \emph{not} for their predictions (as in \selfconsistency{}) but as a means to \emph{collect pieces of evidence} from multiple chains.
Fig.~\ref{fig:multi_chain_example} illustrates \mte{} compared to \selfconsistency{}. 
While both methods rely on sampling multiple reasoning chains, \selfconsistency{} returns the majority answer, \emph{``No''} (grey box, bottom right). By contrast, \mte{} concatenates the intermediate steps from each chain (blue boxes, top left) into a unified context, which is passed, along with the original question, to a \emph{\entailment{}} model. The meta-reasoner is a separate LLM, prompted to meta-reason on multiple reasoning chains and produce a final answer along with an explanation (pink box, bottom left).
By reasoning on multiple reasoning chains, \mte{} is able to mitigate the aforementioned drawbacks -- it combines facts from multiple chains to produce the correct final answer, with an explanation of the answer's validity.




\mte{} has three main components (\S\ref{sec:method}). To generate reasoning chains we use two components, a \emph{\decomposition{}} model and a \emph{\retriever{}} which jointly generate the chain (Fig.~\ref{fig:architecture}), similar to prior work \cite{press2023measuring, trivedi2022interleaving}. These chains are then concatenated into a unified \emph{multi-chain context} which is fed to the aforementioned meta-reasoner.
Fig.~\ref{fig:multi_chain_example} highlights the ability of the meta-reasoner to combine facts from different reasoning chains (intermediate answers in pink). The output explanation combines facts from each of the three chains: (1) \textit{``Seismology is the study of earthquakes''}; (2) \textit{``San Andreas is a film...''}; (3) \textit{``Brad Peyton is a film director, writer...''}. 
SC (in grey) errs due to only using the answers, while the meta-reasoner reads entire reasoning chains, and is able to correctly answer the question.


We evaluate \mte{} on a wide range of challenging multi-hop question answering (QA) datasets, in an open-domain setting.
The datasets can be categorized into two types of tasks: \emph{implicit reasoning} tasks, where reasoning steps are implicit given the question text and
need to be inferred using a strategy \cite{tafjord-etal-2019-quartz, geva-etal-2021-aristotle,kalyan-etal-2021-much}; \emph{explicit reasoning} tasks, where a single reasoning strategy exists and can be directly inferred given the language of the question \cite{yang2018hotpotqa, welbl2018constructing, press2023measuring, Aly2021FEVEROUSFE}. As our baselines, we compare \mte{} to SC, as well as to variants of Self-Ask \cite{press2023measuring} and CoT augmented with retrieval, following \citet{trivedi2022interleaving}. 
Our results show \mte{} consistently outperforms all other baselines, in particular, beating SC by up to 5.7\%, while using the same reasoning chains (\S\ref{sec:experiments}). 

We analyze the qualities of \mte{} in \S\ref{sec:analysis}, by manually scoring its generated explanations and estimating their accuracy. 
Our analysis shows that \mte{} generates high quality explanations for over 82\% of examples, while fewer than 3\% are unhelpful. 
To conclude, our main contributions are:
\begin{itemize}[topsep=0pt, itemsep=0pt, leftmargin=.2in, parsep=0pt]
  \item We introduce the \mte{} method for meta-reasoning on multiple chains-of-thought. 
  \item We show that \mte{} outperforms all baselines, including self-consistency, on all \numdatasets{} multi-hop open-domain QA benchmarks.
  \item We analyze \mte{} for its explanation quality and its multi-chain reasoning capabilities.
\end{itemize}
Our data and codebase are publicly available.\footnote{\url{https://github.com/oriyor/reasoning-on-cots}}

\section{Background}
\label{sec:background}


Recently, there has been a surge of interest in answering multi-hop questions through few-shot prompting of LLMs \cite{Wei2022ChainOT, nye2022show, yao2022react}. The majority of these works follow a common standard:
First, given a question, plan a step-by-step reasoning chain to derive the answer and solve all intermediate steps, aided by a retriever to minimize model hallucination \cite{khot2023decomposed, press2023measuring, yao2022react, lazaridou2023internetaugmented, trivedi2022interleaving, Khattab2022DemonstrateSearchPredictCR}. Then, incorporate multiple reasoning chains with answers to derive the final answer \cite{wang2023selfconsistency,Li2022OnTA}.
In our work, we follow this template and focus on the latter part. However, our \emph{meta-reasoning} approach differs from prior work by reasoning on multiple reasoning chains. Namely, we use multiple chains to collect relevant evidence for question answering.

\begin{figure}[t]\setlength{\belowcaptionskip}{-8pt}
  \centering
  \includegraphics[trim={0cm 1cm 0cm 0cm}, clip, width=0.92\textwidth]{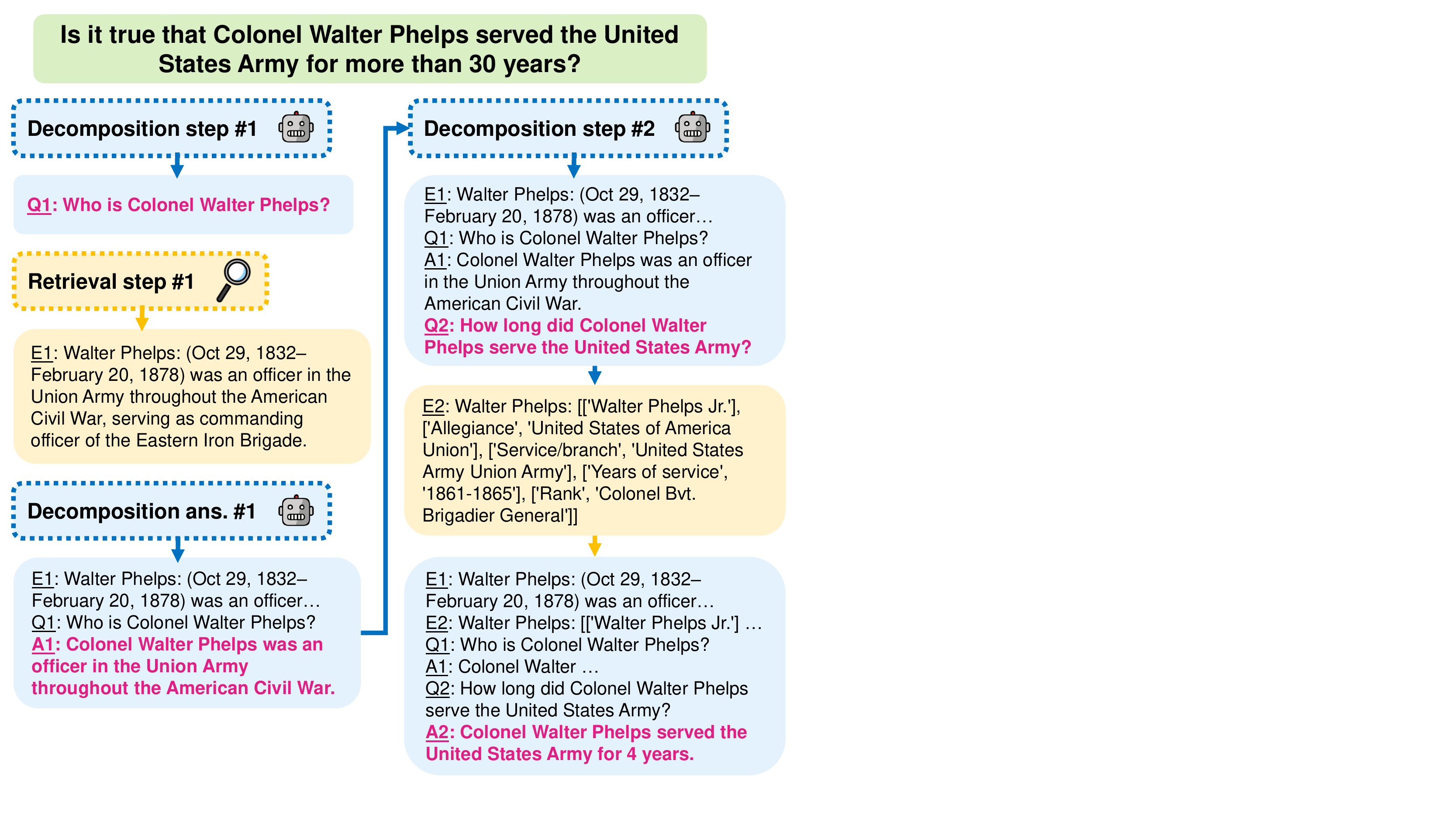}
  \vspace*{-0.7cm}
  \caption{Interleaving decomposition and retrieval steps.}
  \label{fig:decompose_retrieve}
\end{figure}

\section{Method}
\label{sec:method}

We present a method for answering questions by meta-reasoning on multiple reasoning chains. Our focus is on open-domain QA, where the input is a question $q$, and the evidence to answer it is found in one or more sentences in a corpus $C$.
When answering $q$ requires multiple reasoning steps, it can be expressed by a \emph{reasoning chain}, denoted by $r$.
The reasoning chain is a list of one or more intermediate question-evidence-answer triples $(q_i, e_i, a_i)$. Evidence $e_i \in C$ is a sentence that is relevant to answering the intermediate question $q_i$.

Fig.~\ref{fig:architecture} describes our approach when answering  \textit{``How many ants would fit into The Shard?''}.  First, we use a prompted LLM to generate multiple reasoning chains, $r^{(1)},...,r^{(k)}$ (steps 1-2). Each $r^{(j)}$ is generated by interleaving generated intermediate questions with retrieved contexts (\S\ref{sec:method_decomp_retrieve}). Our main contribution is step 3: We introduce a second LLM that is prompted to \emph{meta-reason} on multiple reasoning chains, collecting evidence facts as its explanation and generating the final answer (\S\ref{sec:method_entail}).

\subsection{Generating Reasoning Chains}
\label{sec:method_decomp_retrieve}
Given a question $q$, we generate its reasoning chain using: (1) a \decomposition{} model, and (2) a \retriever{} component. 
Our reasoning chain generation process is largely based on prior work \cite{press2023measuring,trivedi2022interleaving}, discussed in \S\ref{sec:background}. 
Fig.~\ref{fig:decompose_retrieve} describes the interleaving of decomposition and retrieval. 
At each step, the \decomposition{} model generates an  intermediate question $q_i$, based on the original question $q$ and the previous reasoning steps. Then, the \retriever{} uses $q_i$ to retrieve relevant evidence $e_i \in C$. We feed $e_i$ and $q_i$ to the \decomposition{} model (along with the previous steps) to generate intermediate answer $a_i$. During answer generation, we prepend intermediate evidence sentences to the beginning of the chain rather than interleaving them, as it improves the accuracy for all baselines. 
For decomposition prompts, see \S\ref{sec:appendix_prompts}.

\subsection{Reasoning over Reasoning Chains}
\label{sec:method_entail}

The \entailment{} module is the core contribution of \mte{}. Instead of sampling multiple chains for their predicted \emph{answers} \cite{wang2023selfconsistency}, we utilize them for \emph{context generation}. This context is fed to a prompted LLM to \emph{read} the generated chains and \emph{reason} over them to return the answer.

In \S\ref{sec:method_decomp_retrieve}, we defined a reasoning chain as a list of $(q_i, e_i, a_i)$ triples. We first sample multiple chains and use all of their intermediate question-answer pairs $(q_i, a_i)$ as our \emph{multi-chain context} (a variant using question-evidence pairs $(q_i, e_i)$ is described in \S\ref{sec:ent_contexts}). Fig.~\ref{fig:architecture} presents the multi-chain context of the three sampled chains (lower pink box). 
Next, the multi-chain context and the original question are input to the \entailment{}. This model is an LLM, few-shot prompted for QA over a multi-chain context. Fig.~\ref{fig:entailment_prompt} presents one exemplar from the \entailment{} prompt for the \feverous{} dataset (full prompts in \S\ref{sec:appendix_prompts}). We instruct the LLM to \emph{``answer the question step-by-step''} given its multi-chain context, where each line describes a $(q_i, a_i)$ pair from one of the sampled chains. Next, we append the question and a step-by-step reasoning chain followed by the final answer. This last chain serves as the explanation for solving the question.
The \entailment{} is prompted with 6-10 exemplars, based on the dataset (\S\ref{sec:experimentalsetting}). 

Providing the \entailment{} with multiple chains allows it to combine and aggregate facts across chains. Moreover, the model needs to extract the most relevant facts in the chains to serve as its explanation. This enables \mte{} to be both more accurate and more interpretable than past multi-chain approaches (as we analyze in \S\ref{sec:analysis}).

\section{Experiments}
\label{sec:experiments}
We compare \mte{} 
to existing methods on \numdatasets{} multi-hop QA benchmarks. These cover a wide range of reasoning skills, including commonsense, composition, comparison and fact verification.
\mte{} consistently outperforms existing approaches on all benchmarks, when experimenting with two different LLMs and retrievers. Our setting is described in \S\ref{sec:experimentalsetting} and we discuss our main results in \S\ref{sec:results}.

\begin{figure}[t]\setlength{\belowcaptionskip}{-8pt}
\footnotesize
  \begin{tabular}{p{0.97\linewidth}}
    \toprule \bf
    \textit{Given a question and a context, answer the question step-by-step. If you are unsure, answer Unknown.}\\\midrule
    \bf Context:\\
    Who is Robert Broderip? Robert Broderip was an English organist and composer.\\
    Where did Robert Broderip live all his life? Robert Broderip lived in Bristol all his life.\\
    When did Robert Broderip live? Robert Broderip lived during the 19th century.\\
    ...\\
    Where did Robert Broderip live? Broderip lived in Bristol.\\
    During what part of the nineteenth century did Robert Broderip write music? Robert Broderip wrote music during the latter part of the eighteenth century.
    \\\midrule
    \textbf{Question:} Is it true that Robert Broderip lived in London all his life and wrote a considerable quantity of music during the earlier part of the nineteenth century?
 \\\midrule
    \textbf{Answer:} Robert Broderip lived in Bristol all his life, not in London.
    \textbf{So the answer is:} No.\\
     \bottomrule
\end{tabular}
\caption{An exemplar from the \entailment{} prompt.}
\label{fig:entailment_prompt}
\end{figure}

\subsection{Experimental Setting}
\label{sec:experimentalsetting}
\subsubsection{Datasets}
\label{sec:datasets}
As our focus is on multi-hop questions (in an open-domain setting), all datasets require \emph{multiple} reasoning steps. Following prior work \cite{Khattab2022DemonstrateSearchPredictCR, trivedi2022interleaving} and to limit the cost of model API calls, we evaluate on 500-1000 random examples from the development set of each dataset.\footnote{We use the entire development set for \quartz{} and \bamboogle{}, since they include less than 500 examples. For \fermi{} we use all 286 ``Real Fermi Problems'' in its train and development sets. Exact numbers are listed in Tab.~\ref{tab:dev_set_results}.}
We also evaluate on the official test sets of \strategyqa{} and \fermi{}, as they target implicit reasoning, have multiple valid strategies, and their test set evaluation cost is reasonable. For all datasets, we make sure that no evaluation questions appear in any of our prompts. 
Tab.~\ref{tab:datasets} has example questions from each dataset. Our multi-hop QA benchmarks can be categorized based on their required reasoning skills: 

\begin{itemize}[topsep=0pt, itemsep=0pt, leftmargin=.2in, parsep=0pt]
    \item \textbf{Implicit Reasoning:} Questions that entail implicit reasoning steps \cite{geva-etal-2021-aristotle}. The reasoning steps for solving it cannot be explicitly derived from the language of the question and require commonsense or arithmetic reasoning. Such questions may have multiple valid reasoning chains. We evaluate on: \strategyqa{} \cite{geva-etal-2021-aristotle}, \fermi{} \cite{kalyan-etal-2021-much} and \quartz{} \cite{tafjord-etal-2019-quartz}.
    \item \textbf{Explicit Reasoning:}
    Multi-hop questions where the reasoning steps are explicitly expressed in the language of the question (composition, comparison). 
    These include \hotpotqa{} \cite{yang2018hotpotqa}, \wikihop{} \cite{welbl2018constructing} and \bamboogle{} \cite{press2023measuring}. We also evaluate on \feverous{} \cite{Aly2021FEVEROUSFE}, a fact verification dataset where claims require 
    verifying multiple facts, and evidence may be either in sentences, tables or both.
\end{itemize}

For evaluation, we use \fone{} to compare predicted and gold answers for all explicit reasoning datasets and exact-match for the binary-choice datasets. In \fermi{}, we use the official order-of-magnitude evaluation by \citet{kalyan-etal-2021-much}. We provide additional technical details on evaluation in \S\ref{sec:evaluation_appendix}.

\begin{table}[t!]
\begin{center}
\footnotesize
\begin{tabular}{p{1.8cm}p{5cm}}\toprule
\bf Dataset & \bf Example  \\ \midrule
\strategyqa{} (implicit) & Can Arnold Schwarzenegger deadlift an adult Black rhinoceros?
\\ \midrule
\fermi{} \hspace{5mm}(implicit) & How many high fives has LeBron James given/received? 
\\ \midrule
\quartz{} \hspace{5mm}(implicit) & Jeff drained his rice field in the wintertime. The field likely will produce \_\_ crops when he uses it. A. more B. less
\\ \midrule
\hotpotqa{} (explicit) & What city did the musician whose debut album shares its title with the 1959 Alfred Hitchcock film hail from? 
\\ \midrule
\wikihop{} (explicit) & Where was the place of death of Isabella of Bourbon's father? 
\\ \midrule
\bamboogle{} (explicit) & What is the maximum airspeed (in km/h) of the third fastest bird? 
\\ \midrule
\feverous{} (explicit) & Is it true that Robert Broderip lived in London all his life and wrote a considerable quantity of music during the earlier part of the nineteenth century? 
\\ 
\bottomrule
\end{tabular}
\end{center}
\caption{The multi-hop QA datasets in our experiments.}
\label{tab:datasets}
\end{table}

\begin{table*}[t]
\centering
\footnotesize
  \begin{tabular}{ccccccccc}
    \toprule
    Dataset & Reasoning & Examples & Oracle & \selfask & \selfconsistency{}$@3$  & \selfconsistency{}$@5$  & \ste & \mte \\
    \midrule

    \strategyqa &  & 1,000 & 94.4$\pm$0.1 & 69.3$\pm$0.3 & 71.5$\pm$0.8 & 72.2$\pm$0.8 & 70.0$\pm$0.6 & \textbf{73.6$\pm$0.7}\\
    \fermi & implicit & 286 & 65.1$\pm$0.8 & 38.3$\pm$0.7 & 38.4$\pm$0.7 & 38.3$\pm$0.8 & 38.1$\pm$0.8 & \textbf{38.9$\pm$0.8}\\
    \quartz &  & 374 & 94.1$\pm$0.5 & 78.3$\pm$0.4 & 78.2$\pm$0.7 & 77.6$\pm$0.5 & 80.7$\pm$0.1 & \textbf{81.6$\pm$1.3}\\
    \midrule
    \hotpotqa &  & 500 & 68.0$\pm$0.4 & 50.2$\pm$0.3 & 50.5$\pm$0.8 & 51.3$\pm$0.2 & 56.4$\pm$0.4 & \textbf{57.0$\pm$0.8}\\
    \wikihop & explicit & 500 & 77.5$\pm$0.8 & 63.8$\pm$0.1 & 64.5$\pm$0.8 & 65.4$\pm$0.6 & 67.2$\pm$0.2 & \textbf{67.9$\pm$0.4}\\
    \bamboogle &  & 120 & 77.3$\pm$0.5 & 64.6$\pm$0.6 & 64.6$\pm$0.4 & 65.0$\pm$1.5 & 64.7$\pm$0.4 & \textbf{66.5$\pm$1.7}\\
    \feverous &  & 500 & 88.0$\pm$0.4 & 66.0$\pm$1.0  & 67.8$\pm$0.2 & 67.9$\pm$0.6 & 65.1$\pm$0.4  & \bf69.4$\pm$1.0 \\
  \bottomrule
\end{tabular}
\caption{Experiments using \codex{} on seven multi-hop open-domain QA datasets. Results are averaged over 3 runs. \bamboogle{} results are averaged over 5 runs due to its smaller size.}
  \label{tab:dev_set_results}
\end{table*}

\subsubsection{Models}
\label{sec:models}
Our main models and baselines are all retrieval-augmented instances of \codex{}, prompted with in-context learning exemplars \cite{NEURIPS2020_1457c0d6}. In \S\ref{sec:colbert_vicuna}, we include additional experiments with the open-source \vicuna{} \cite{vicuna2023} LLM. Prompt exemplars are formatted as described in \S\ref{sec:method_entail}. The number of exemplars varies from 6-12 between datasets.
Decomposition prompt exemplars are based on random examples from the train and development sets, coupled with their gold reasoning chain. 
For the \entailment{} exemplars, we use reasoning chains sampled from the \decomposition{} model as the multi-chain context.
We ensure that the answer can be inferred using the sampled chains and add an explanation before the final answer, as shown in Fig.~\ref{fig:entailment_prompt}.
For the binary-choice datasets, \strategyqa{}, \quartz{}, and \feverous{}, the prompt contains an equal number of exemplars from each label. For additional details regarding the full prompts, length statistics and robustness to a different choice of prompts, please refer to \S\ref{sec:appendix_prompts}.

\paragraph{Meta-Reasoner} 
We experiment with two variants of the \entailment{} to measure the effect of reasoning on more than a single chain.

\begin{itemize}[topsep=0pt, itemsep=0pt, leftmargin=.2in, parsep=0pt]
    \item \textbf{\mte{}:} 
    The meta-reasoner is given five reasoning chains as its multi-chain context (\S\ref{sec:method_entail}). We decode one chain with greedy decoding, and sample another four reasoning chains with temperature $t=0.7$.\footnote{Like \citet{wang2023selfconsistency}, we observe that greedy-decoded chains have higher accuracy compared to the other chains.}
    This enables the meta-reasoner to review different pieces of evidence when answering the full question (\S\ref{sec:analysis}).

    \item \textbf{\ste{}:} Single-Chain Reasoning (\ste{}) 
    serves as an ablation for the effect of the multi-chain context. In \ste{}, the meta-reasoner is given the same prompt as \mte{} aside from having only the greedy-decoded chain in its context.
    This disentangles the effect of using multiple chains from the effect of having an LLM that is separate from the decomposition model to generate the final answer.
\end{itemize}



\paragraph{Baselines} We evaluate the following baselines: 
\begin{itemize}[topsep=0pt, itemsep=0pt, leftmargin=.2in, parsep=0pt]
    \item \textbf{\selfask{}:} Self-Ask \cite{press2023measuring} returns the answer of a single reasoning chain, that was generated with greedy decoding.
    \item \textbf{\selfconsistency{}:} Self-Consistency serves as a baseline which incorporates multiple reasoning chains \cite{wang2023selfconsistency}. It returns the majority answer based on multiple chains sampled from the \decomposition{} model. We experiment with variants of 3, 5 and 15 sampled chains (\selfconsistency{}$@3$, \selfconsistency{}$@5$ and \selfconsistency{}$@15$), in line with prior work \cite{wang2023selfconsistency, Khattab2022DemonstrateSearchPredictCR,sun2022recitation}. As in \mte{}, we use the chain generated with greedy decoding along with additional chains sampled with $t=0.7$.
\end{itemize} 


\paragraph{Retrieval} Similar to \citet{press2023measuring, lazaridou2023internetaugmented,Paranjape2023ARTAM}, our models and baselines use a \retriever{} based on Google Search, via the SerpAPI service.\footnote{\url{https://serpapi.com/}} 
However, we also include results using an open-source retriever \cite{khattab2020colbert} in \S\ref{sec:colbert_vicuna}.
As most of our datasets contain evidence from Wikipedia (\S\ref{sec:datasets}), we consider it as our retrieval corpus. Therefore, we format search queries as ``\texttt{en.wikipedia.org $q_i$}'', with the Wikipedia domain preceding the intermediate question. We return the top-1 evidence retrieved by Google. Retrieved evidence may be either sentences or parsed lists.
Following \citet{trivedi2022interleaving}, we also retrieve evidence for the original question $q$. Last, all retrieved evidence sentences are
prepended to the decomposition (\S\ref{sec:method_decomp_retrieve}). Additional implementation details about our retrieval and \mte{} are described in \S\ref{sec:retrieval} and \S\ref{sec:hps}.

\subsection{Main Results}
\label{sec:results}
Next, we report our evaluation results. Overall, \mte{} outperforms our baselines on all \numdatasets{} datasets.

\paragraph{\mte{} Performance} Tab.~\ref{tab:dev_set_results} presents the results for all \numdatasets{} multi-hop datasets (evaluation described in \S\ref{sec:datasets}). We evaluate both \selfconsistency{}$@5$ and \mte{} using five reasoning chains. In addition, we list an \emph{oracle score} which uses the best answer out of all five chains.
\mte{} outperforms all baselines on all of the benchmarks, beating \selfconsistency{}$@5$ on  \strategyqa{} (+1.4\%), \fermi{} (+0.6\%), \quartz{} (+4.0\%), \hotpotqa{} (+5.7\%), \wikihop{} (+2.5\%), \bamboogle{} (+1.5\%) and \feverous{} (+1.5\%).

\begin{figure}[t]
  \centering
  \includegraphics[trim={0cm 0cm 0cm 0cm}, clip, width=\columnwidth]{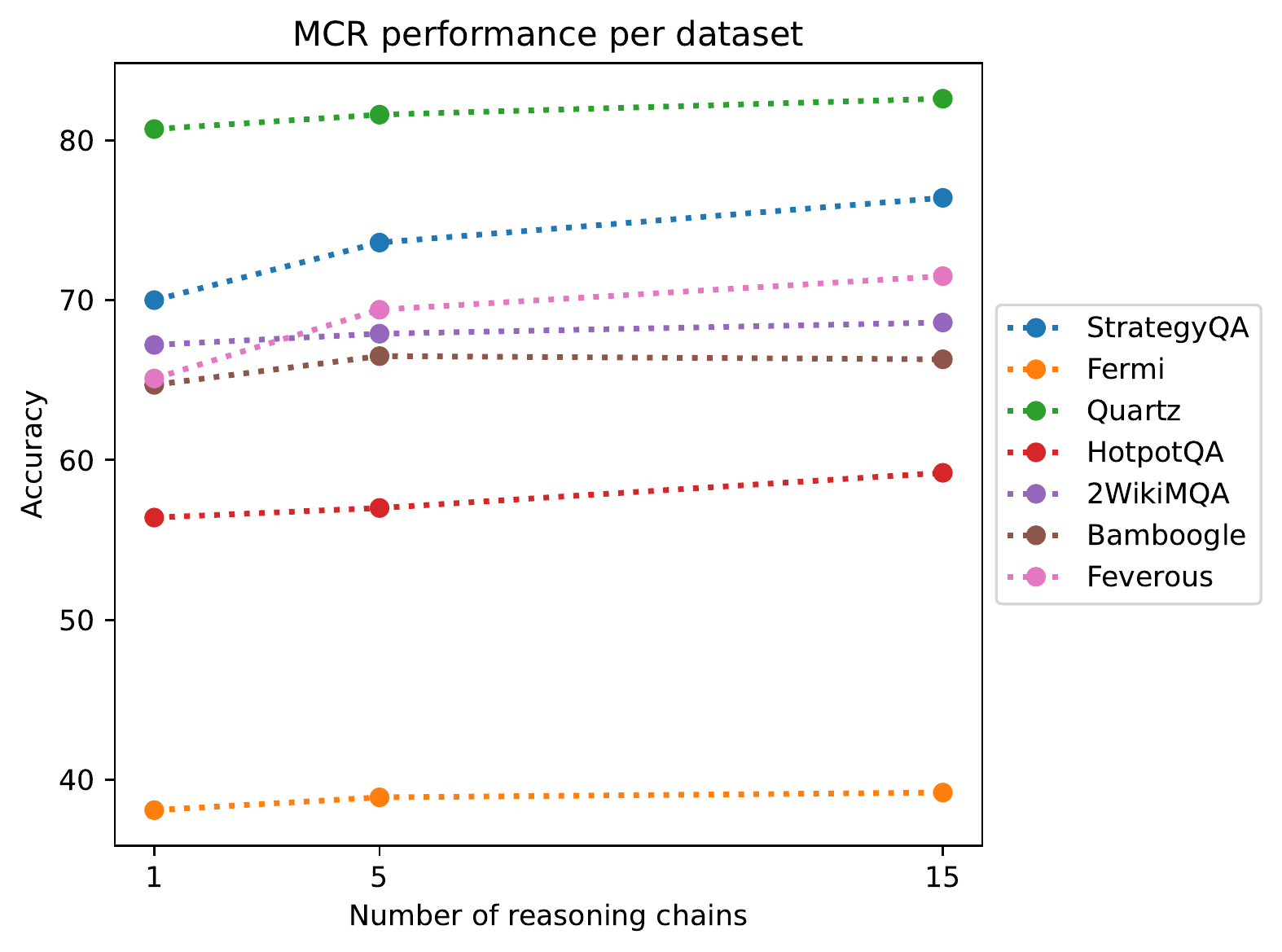}
  \caption{\label{fig:mcr_num_chains_plot}Per-dataset performance as a function of the number of reasoning chains used by \mte{} (1, 5, 15).}
\end{figure}

\paragraph{Adding Reasoning Chains} 
We measure the gains of \mte{} and \selfconsistency{} when adding reasoning chains. As extending \mte{} is bounded by context length,\footnote{\codex{} context is capped at 8,001 tokens.} we follow a straightforward approach and perform self-consistency on three \mte{} runs.
We compare this model, \mte$+$\selfconsistency{}$@3$, which used 15 reasoning chains (5 for each \mte{} run), to \selfconsistency{}$@15$.
Tab.~\ref{tab:mte-sc} shows that \mte$+$\selfconsistency{}$@3$ consistently outperforms \selfconsistency{}$@15$. Furthermore, though \mte{} uses only 5 reasoning chains, it beats \selfconsistency{}$@15$ on all datasets, save \strategyqa{}. Fig.~\ref{fig:mcr_num_chains_plot} plots, for each dataset, the effect that adding more reasoning chains has on meta-reasoning performance. It presents the results with 1 chain (\ste{}), 5 chains (\mte{}) and 15 reasoning chains (\mte$+$\selfconsistency{}$@3$).

\begin{table}[t]
\centering
\footnotesize
  \begin{tabular}{cccccc}
    \toprule
    Dataset  & \selfconsistency{}$@15$   & \mte{} & \mte$+$\selfconsistency{}$@3$ \\
    \midrule

    \strategyqa & 74.6 & 73.6 & \bf76.4\\
    \fermi & 38.6 & 38.9 & \bf39.2\\
    \quartz & 78.3 & 81.6 & \bf82.6\\
    \midrule
    \hotpotqa & 54.1 & 57.0 & \bf59.2\\
    \wikihop & 65.8 & 67.9 & \bf68.6 \\
    \bamboogle & 65.6 & \bf66.5 & 66.3 \\
    \feverous  & 68.6 & 69.4 & \bf71.5\\
  \bottomrule
\end{tabular}
\caption{Running \selfconsistency{} and \mte{} on 15 reasoning chains.}
  \label{tab:mte-sc}
\end{table}

\paragraph{Test Set Results} We evaluate our models on the official test sets of \strategyqa{}\footnote{\url{https://leaderboard.allenai.org/strategyqa}} and \fermi{}, which include 490 and 558 examples respectively. The results in Tab.~\ref{tab:test_sets} show that on \strategyqa{} \mte{} consistently beats \selfconsistency{}, when using the same number of reasoning chains. 
In \fermi{}, both methods perform similarly.

\begin{table}[t]
\centering
\footnotesize
  \begin{tabular}{lccc}
    \toprule
     Model & \# chains & \strategyqa  & \fermi \\
     \midrule
     \selfconsistency{}$@5$ & 5 & 71.4 & \bf39.8 \\
     \mte{} & 5 & \bf72.5 &  39.7 \\\midrule
     \selfconsistency{}$@15$ & 15 & 74.1 & 39.7\\
     \mte$+$\selfconsistency{}$@3$ & 15 & \bf75.3 & \bf 40.1 \\
  \bottomrule
\end{tabular}
\caption{Test set results for \strategyqa{} and \fermi{}. }
  \label{tab:test_sets}
\end{table}

\paragraph{Recent Approaches}
Previously, we established the advantages of meta-reasoning over multiple reasoning chains.
While an apples-to-apples comparison with other recent approaches is impossible due to fundamental differences in the experimental setup (see \S\ref{sec:recent_approaches_comparison}), it serves as a rough measuring stick for the robustness of \mte{} across different tasks.
In \S\ref{sec:recent_approaches_comparison}, Tab.~\ref{tab:past_work} we compare \mte{} to five recent CoT-based approaches for multi-hop QA. \mte{} performance is comparable with the best results on all datasets (shared between these works), showcasing its robustness.

\subsection{Open-source Models}
\label{sec:colbert_vicuna}

To further examine \mte{}'s performance (\S\ref{sec:results}) and for better reproducibility, we experiment with an additional open-source retriever and LLM. As our retriever, we use ColBERTv2 \cite{santhanam-etal-2022-colbertv2} over the 2018 Wikipedia dump from \citet{karpukhin-etal-2020-dense}. 
In addition to \codex{}, we experiment with \vicuna{} \cite{vicuna2023}, a 13-billion parameters model shown to outperform LLMs like LLaMA and Alpaca \cite{Touvron2023LLaMAOA, alpaca}. We use the same prompts as in \codex{}, trimmed to fit a 2,048 tokens context length. 

We report the full results of the open-source ColBERTv2 retriever with \codex{} and \vicuna{} in Tab.~\ref{tab:vicuna_colbert_results_appendix}. In addition, we provide results of open-source models when reasoning over 15 reasoning chains in Tab.~\ref{tab:mte-sc_colbert_appendix}.
For \codex{}, substituting Google Search with ColBERTv2 exhibits the same trend as in Tab.~\ref{tab:dev_set_results}, albeit a slight decrease in performance. \mte{} outperforms all other baselines, beating \selfconsistency{}$@5$ on \strategyqa{} (+2.3\%), \fermi{} (+3.4\%), \quartz{} (+3.9\%), \hotpotqa{} (+3.5\%), \wikihop{} (+1.2\%), \bamboogle{} (+3.6\%) and \feverous{} (+1.4\%).
Unsurprisingly, results sharply decrease when evaluating the smaller \vicuna{} with ColBERTv2.
The comparison between \mte{} and \ste{} suggests that reasoning over multiple chains is a challenge for the weaker \vicuna{} model. For example, it generates open-ended answers such as \textit{``Unknown''} or \textit{``It depends''} for over $24\%$ of the questions in \strategyqa{}. This suggests that meta-reasoning over multiple chains has greater gains (compared to \ste{}) when both the decomposition model and meta-reasoner are larger LLMs.

However, even on \vicuna{}, \mte{} still outperforms all baselines on 5 datasets and beats \selfconsistency{}$@5$ on all 7 of them: \strategyqa{} (+0.5\%), \fermi{} (+4.6\%), \quartz{} (+3.6\%), \hotpotqa{} (+6.5\%), \wikihop{} (+0.3\%), \bamboogle{} (+3.0\%) and \feverous{} (+1.3\%). When evaluating with 15 reasoning chains, in Tab.~\ref{tab:mte-sc_colbert_appendix}, \mte$+$\selfconsistency{}$@3$ continually beats \selfconsistency{}$@15$.

\begin{table*}[t]
\centering
\footnotesize
  \begin{tabular}{cccccccc}
    \toprule
    Dataset & Model & Oracle & \selfask & \selfconsistency{}$@3$  & \selfconsistency{}$@5$  & \ste & \mte \\
    \midrule

    \strategyqa & \codex{} & 94.5$\pm$0.7 & 67.1$\pm$0.6 & 69.9$\pm$0.1 & 70.8$\pm$0.6 & 67.8$\pm$0.5 & \textbf{73.1$\pm$2.1}\\
    \fermi & \codex{} & 64.3$\pm$0.7 & 33.2$\pm$0.3 & 33.2$\pm$0.4 & 33.1$\pm$0.4 & 33.9$\pm$0.6 & \textbf{36.5$\pm$2.1} \\
    \quartz &  \codex{} & 93.9$\pm$0.6 & 77.1$\pm$0.6 & 75.6$\pm$0.7 & 76.0$\pm$1.5 & 79.3$\pm$0.3 & \textbf{79.9$\pm$1.2}\\
    \hotpotqa & \codex{} & 67.7$\pm$0.7 & 50.7$\pm$0.3 & 51.5$\pm$0.6 & 52.5$\pm$0.1 & 55.3$\pm$0.2 & \textbf{56.0$\pm$1.1}\\
    \wikihop & \codex{} & 68.3$\pm$0.4 & 52.4$\pm$0.1 & 51.1$\pm$0.2 & 52.7$\pm$0.4 & 53.7$\pm$0.3 & \textbf{53.9$\pm$0.3}\\
    \bamboogle & \codex{} & 56.4$\pm$1.2 & 45.9$\pm$1.1 & 47.2$\pm$1.4 & 47.0$\pm$0.7 & 47.1$\pm$1.0 & \textbf{50.6$\pm$1.3} \\
    \feverous & \codex{} & 84.1$\pm$0.7 & 61.2$\pm$0.4 & 
    62.9$\pm$0.6 & 63.1$\pm$1.0 & 60.9$\pm$0.3  & \textbf{64.5$\pm$0.8} \\
    \midrule
    \strategyqa & \vicuna{} & 82.4$\pm$0.2 & 59.7$\pm$0.1 & 61.4$\pm$0.5 & 62.2$\pm$0.8 & \textbf{63.7$\pm$0.0} & 62.7$\pm$0.1\\
    \fermi & \vicuna{} & 45.7$\pm$1.0 & 19.1$\pm$0.2 & 19.1$\pm$0.3 & 18.8$\pm$0.3 & 21.5$\pm$0.0 & \textbf{23.4$\pm$0.4} \\
    \quartz & \vicuna{} & 89.6$\pm$1.6 & 61.1$\pm$0.1 & 59.8$\pm$2.3 & 61.4$\pm$1.6 & 63.9$\pm$0.0 & \textbf{65.0$\pm$0.3} \\
    \hotpotqa & \vicuna{} & 52.7$\pm$0.5 & 34.8$\pm$0.0 & 35.8$\pm$0.2 & 37.1$\pm$0.4 & 43.4$\pm$0.0 & \textbf{43.6$\pm$1.6}\\
    \wikihop & \vicuna{} & 52.2$\pm$0.3 & 32.2$\pm$0.4 & 33.8$\pm$0.6 & 34.0$\pm$1.0 & \textbf{35.1$\pm$0.0} & 34.3$\pm$0.4\\
    \bamboogle & \vicuna{} & 42.3$\pm$1.6 & 30.7$\pm$0.0 & 30.4$\pm$0.6 & 31.4$\pm$0.6 & 31.3$\pm$0.0 & \textbf{34.4$\pm$1.3}\\
   \feverous & \vicuna{} & 88.7$\pm$0.2 & 61.5$\pm$0.6 & 61.0$\pm$0.6 & 61.0$\pm$0.8 & 60.6$\pm$0.0 & \textbf{62.3$\pm$1.2} \\ 
  \bottomrule
\end{tabular}
\caption{Experiments using ColBERTv2 retriever with \codex{} and \vicuna{}, evaluated on the development sets of our datasets. Results are averaged over 3 runs. The number of examples per dataset is the same as in Tab.~\ref{tab:dev_set_results}. \vicuna{} generation is deterministic so, using greedy decoding, \ste{} has a standard deviation of $0$.}
  \label{tab:vicuna_colbert_results_appendix}
\end{table*}

\begin{table*}[t]
\centering
\footnotesize
  \begin{tabular}{c|ccc|ccc}
    \toprule
    Dataset & Model & \selfconsistency{}$@15$  & \mte$+$\selfconsistency{}$@3$ & Model & \selfconsistency{}$@15$  & \mte$+$\selfconsistency{}$@3$ \\
    \midrule
    \strategyqa & \codex{} & 72.6 & \bf75.6 & \vicuna{} & 62.3 & \bf63.7\\
    \fermi & \codex{} & 34.0 & \bf36.3 & \vicuna{} & 18.8 & \bf23.2\\
    \quartz & \codex{} & 76.5 & \bf80.7 & \vicuna{} & 60.1 & \bf64.3\\
    \hotpotqa & \codex{} & 54.3 & \bf56.8 & \vicuna{} & 37.8 & \bf44.8\\
    \wikihop & \codex{} & 52.5 & \bf54.0 & \vicuna{} & 35.5 & \bf35.6\\
    \bamboogle & \codex{} & 48.9 & \bf51.8 & \vicuna{} & 31.8 & \bf35.1\\
    \feverous & \codex{} & 62.7 & \bf66.2 & \vicuna{} & 61.1 & \bf64.0\\ 
  \bottomrule
\end{tabular}
\caption{Running \selfconsistency{} and \mte{} on 15 reasoning chains using ColBERTv2 retriever with \codex{} (left columns) and \vicuna{} (right columns).}
  \label{tab:mte-sc_colbert_appendix}
\end{table*}
\section{Analysis}
\label{sec:analysis}

Next, we measure the importance of incorporating multiple reasoning chains in \mte{} and qualitatively assess its output.

\begin{figure}[t]\setlength{\belowcaptionskip}{-8pt}
  \centering
  \includegraphics[trim={0cm 3.5cm 13cm 0cm}, clip, width=0.99\textwidth]{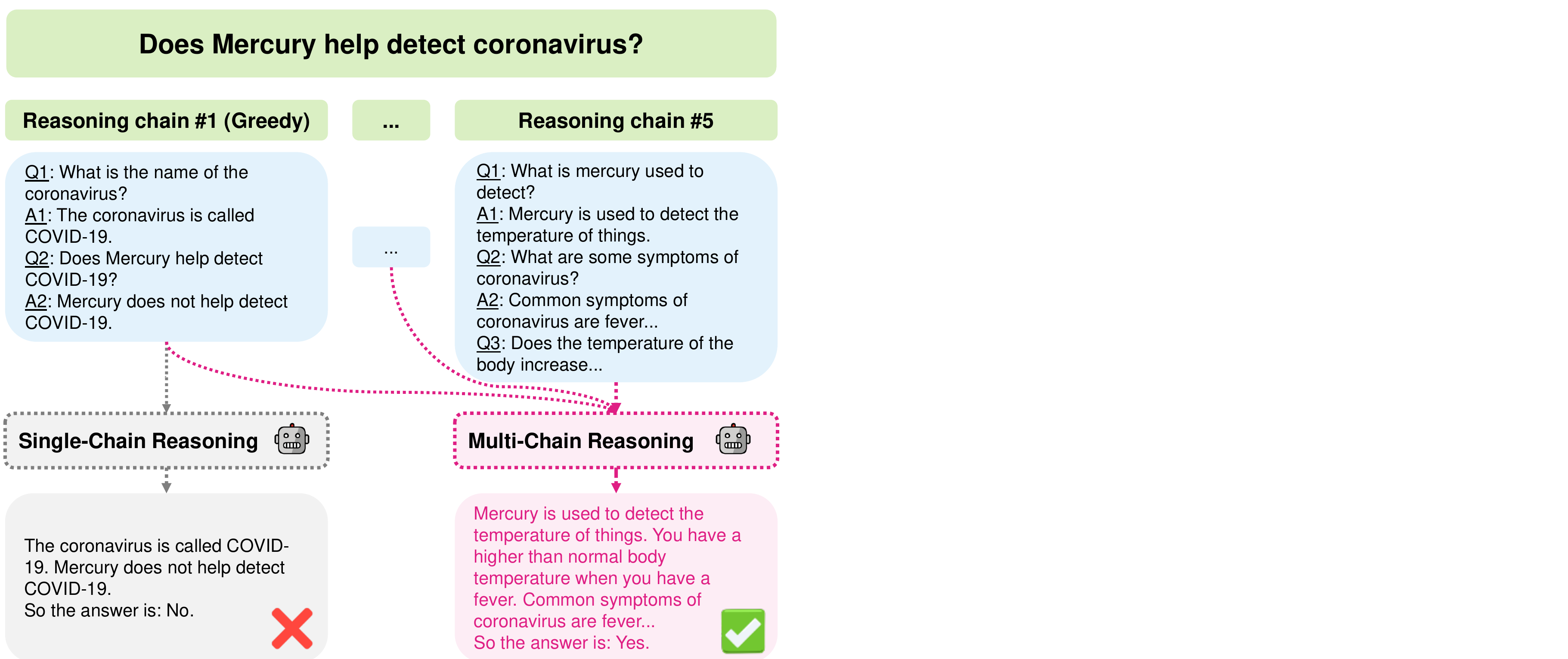}
  \caption{An example from \strategyqa{} where the greedy chain is insufficient to answer the question. \mte{} beats \ste{} by having access to multiple chains.}
  \label{fig:mte_vs_ste_ex}
\end{figure}

\begin{figure}[t]\setlength{\belowcaptionskip}{-8pt}
  \centering
  \includegraphics[ width=0.48\textwidth]{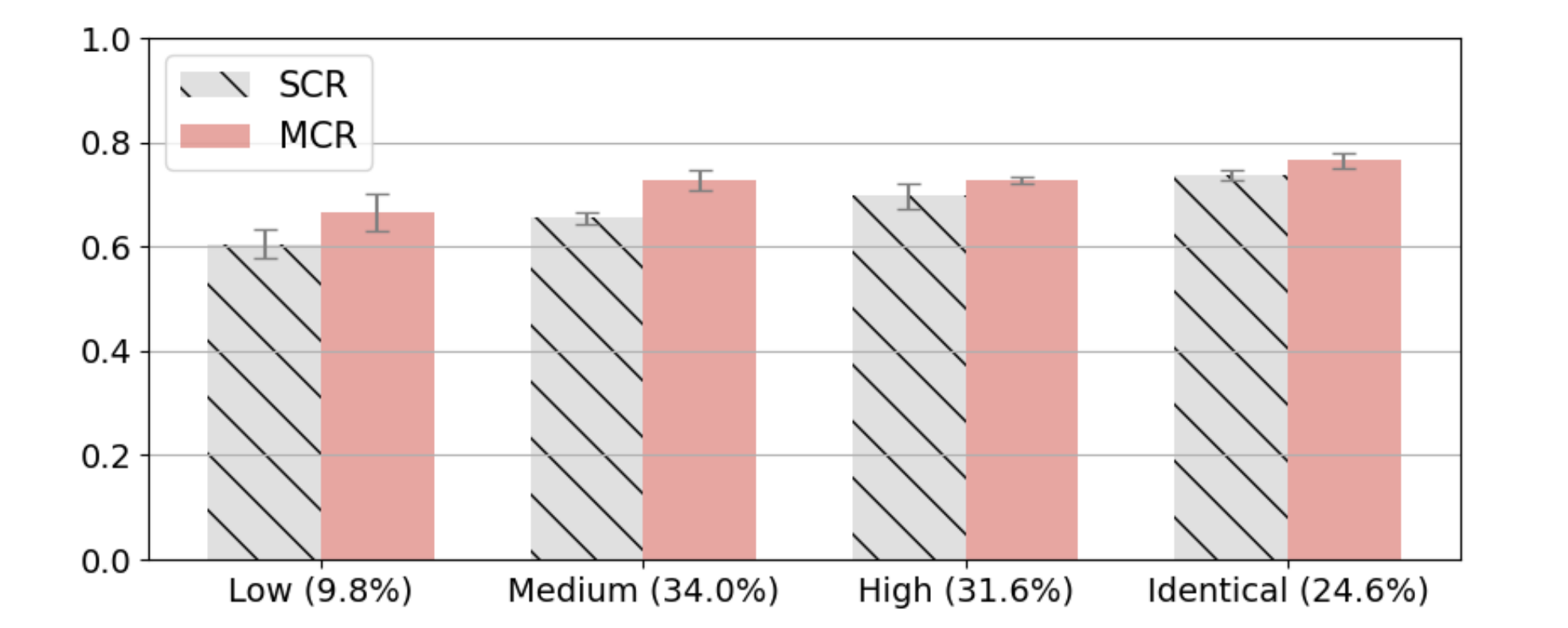}
  \vspace*{-0.7cm}
  \caption{\mte{} and \ste{} accuracy on \strategyqa{}, categorized by the similarity of the greedy  chain to the \mte{} explanation. When \mte{} uses a chain other than the greedy one (lower similarity), it outperforms \ste{}.}
  \label{fig:sim_to_greedy}
\end{figure}

\paragraph{When are Multiple Chains Helpful?}
In \S\ref{sec:results} we observed that \mte{} consistently outperforms single-chain reasoning (\ste{}). We wish to prove that this advantage lies in cases where the meta-reasoner uses additional chains. 
To this end, we sort examples based on the similarity of their greedy-decoded chain to the \mte{} explanation (details in \S\ref{sec:appendix_multiple_chains}). 
Lower similarity indicates less reliance of \mte{} on the greedy chain. 
Fig.~\ref{fig:mte_vs_ste_ex} presents an example where the \mte{} explanation (pink box) includes relevant facts from a chain other than the greedy one (additional examples in \S\ref{sec:combining_chains}). 
Results in Fig.~\ref{fig:sim_to_greedy} empirically demonstrate that on \strategyqa{}, \mte{} gains over \ste{} are highest when \mte{} explanations are less similar to the greedy chain. 
We observe this trend in all datasets (\S\ref{sec:appendix_multiple_chains}), serving as further evidence for \mte{}'s strengths. 



\paragraph{Combining Reasoning Chains}
In addition to choosing between reasoning chains, an interesting property of the meta-reasoner is that it can \emph{combine} facts from different chains.
We estimate the prevalence of this phenomenon on the implicit datasets, \strategyqa{} and \fermi{}, which are more challenging.
Given an example, we automatically check if its \entailment{} explanation is the result of combining chains. 
We examine if one of the output sentences appears in exactly one chain, while another sentence is absent from that chain and is part of a different chain. 
We consider sentences as similar if their ROUGE-1 precision is above 0.8, and distinct if it is below 0.2.
Overall, in 20\% of \strategyqa{} examples and 25\%\ of \fermi{}, the \mte{} explanation results from combining reasoning chains.
From a manual analysis of 50 such examples for each dataset, we observe that these multi-chain explanations are better than any individual reasoning chain in $10\%$ of cases (see examples in \S\ref{sec:combining_chains}, Fig.~\ref{fig:mixing_traces_examples}).
For the remaining $90\%$, the reasoning expressed in the resulting combination is a paraphrase of an individual chain.

\paragraph{Explanation Quality}
The \entailment{} is prompted to generate an explanation alongside the final answer (\S\ref{sec:method_entail}). Inspired by past work \cite{pruthi2021evaluating}, we test the quality of the \mte{} explanations. Four of the authors manually reviewed 600 random examples, 100 per dataset (sans \feverous{} \S\ref{sec:hps}) and scored their meta-reasoner explanations. Each explanation is scored as either 1 (irrelevant), 2 (partially relevant) or 3 (highly relevant), based on its relevance to answering the question.
We find the explanation is highly relevant in 82\% of the cases (87\% excluding \fermi{}, which is the most challenging), and is irrelevant in less than 3\%.  

Next, we evaluate the \emph{faithfulness} of explanations \cite{jacovi-goldberg-2020-towards}, namely, whether a person provided only with the question and \mte{} explanation would answer the same as the model. Our focus was on examples with quality explanations (score 3), since they are answerable given the explanation. We answered each question based on the model's explanation. In 90\% of cases (95\% excluding \fermi{}), the \mte{} predictions matched our own, highlighting the faithfulness of its explanations.
We attribute part of the gap between human and MCR predictions to implicit reasoning tasks, where humans lead by five points, on average. For the full results, see \S\ref{sec:explanation_quality_analysis}.

\paragraph{Error Analysis}
We manually analyzed 700 errors by \mte{} (100 per dataset). 
We consider the following categories: \emph{Valid} predictions where the generated answer is accurate or the original question is ambiguous; \emph{Decomposition} errors where no chain has the necessary reasoning steps to answer the question; \emph{Retrieval} errors where the retrieved contexts were irrelevant, leading the model to hallucinate; \emph{Explanation} errors where \mte{} generates a wrong explanation while a correct one is present in the multi-chain context; \emph{Answer} errors are when the \mte{} explanation is correct, but the answer is not; \emph{Contradicting} facts are cases where \mte{} errs due to contrasting statements appearing in the multi-chain context.

Tab.~\ref{tab:error_analysis} lists the prevalence of the error categories per dataset.
In four datasets, over 20\% of errors appear to be valid predictions, labeled as incorrect due to ambiguous questions, outdated answers or dataset errors.
Decomposition is a challenge in the implicit datasets, \strategyqa{} and \fermi{}, with more than 24\% of errors. Comparing errors on different reasoning datasets (excluding valid examples): Explanation and Answer
errors are 50\% on implicit reasoning datasets compared to 23\% on explicit reasoning ones; Retrieval errors are more prevalent in explicit reasoning tasks with 66\% of errors being due to Retrieval or Contradicting facts, compared to 30\% in implicit datasets.
Additional technical details on our analysis are in \S\ref{sec:error_analysis}.

\begin{table}[t]
\centering
\scriptsize
  \begin{tabular}{c|cccccc}
    \toprule
    Dataset & Va. & De. & Re. & Co. & Ex. & An. \\
    \midrule
    \strategyqa & 20\%  & 24\%  & 8\%  & 15\%  & 20\%  & 17\% \\
    \fermi & 6\%  & 39\%  & 20\%  & 4\%  & 17\%  & 23\% \\
    \quartz & 14\%  & 6\%  & 13\%  & 19\%  & 11\%  & 40\%\\
    \midrule
    \hotpotqa & 33\%  & 24\%  & 24\%  & 11\%  & 11\%  & 5\% \\
    \wikihop & 39\%  & 4\%  & 35\%  & 12\%  & 8\%  & 6\% \\
    \bamboogle & 26\%  & 8\%  & 32\%  & 24\%  & 13\%  & 0\%  \\
    \feverous & 7\%  & 14\%  & 34\%  & 23\%  & 20\%  & 6\% \\
  \bottomrule
\end{tabular}
\caption{Error classes per dataset: Valid (Va.), Decomposition (De.), Retrieval (Re.), Contradicting facts (Co.), Explanation (Ex.) and Answer (An.). We allow multiple error categories per example.}
  \label{tab:error_analysis}
\end{table}
\section{Related Work}
\label{sec:related_work}
 
For a thorough survey on LLM reasoning see \citet{lu2022survey, huang2022reasoning, qiao2022reasoning}. 
A slew of recent works have focused on eliciting multi-step reasoning in LLMs, including scratchpads \cite{nye2022show}, chain-of-thought prompting \cite{Wei2022ChainOT, Zhou2022LeasttoMostPE}, learned verifiers \cite{cobbe2021training}, selection-inference
\cite{Creswell2022SelectionInferenceEL} and bootstrapping \cite{zelikman2022star}.

Self-consistency \cite{wang2023selfconsistency, Fu2022ComplexityBasedPF} selects the majority answer across multiple chains, outperforming learned verifiers and ``sample-and-rank''  approaches \cite{adiwardana2020humanlike, DeFreitas2020TowardsAH}. \newcite{Li2022OnTA} further improve SC by increasing chains' diversity and introducing a trained verifier. \citet{tafjord2022entailer} over-samples chains and verifies them using a natural language inference model on intermediate steps, while \newcite{he2022rethinking} re-rank chains based on intermediate retrieved evidence.
In addition, meta-reasoning is closely tied to \emph{self-reflection} in LLMs, which is becoming increasingly important in using the LLM to review multiple strategies \cite{yao2023tree, shinn2023reflexion, madaan2023selfrefine}.

Recent works proposed revising LLM-generated texts by using retrieved sentences \cite{gao2022rarr} or model-generated feedback \cite{madaan2023selfrefine, chen2023teaching, paul2023refiner}. \mte{} similarly reviews LLM-generated reasoning chains however, its focus is meta-reasoning on \emph{multiple} chains. 

Significant QA research has been dedicated to reasoning over multiple facts retrieved from an underlying corpus. Such tasks include multi-step questions that require explicit reasoning \cite{talmor2018web, welbl2018constructing, wolfson2020break, trivedi2021musique}, implicit reasoning \cite{geva-etal-2021-aristotle} and multi-modal capabilities \cite{talmor2021multimodalqa}.

Recent works also target retrieval-augmented LLMs, prompted to solve open-domain questions  \cite{ lazaridou2023internetaugmented,Khattab2022DemonstrateSearchPredictCR, trivedi2022interleaving,ram2023ralm, yoran2023making}.

\section{Conclusion}
This work introduces \mte{} for meta-reasoning over multiple reasoning chains. We evaluate \mte{} on \numdatasets{} datasets for multi-hop QA that require both implicit and explicit reasoning in an open-domain setting and show that it outperforms previous approaches on all evaluation benchmarks. 

\section{Limitations}
\label{sec:limitations}

In this work we introduce a meta-reasoner model to reason over multiple reasoning chains. While we opt for a prompted LLM as our meta-reasoner, we do not experiment with a fine-tuned meta-reasoning model. 
For the \entailment{} context, we experiment with variants which include either generated QA pairs or retrieved evidence sentences. We leave further improvements to the meta-reasoner context as future work. 
Due to the inference costs of current state-of-the-art LLMs we evaluate on the \codex{} model, similar to prior work \cite{trivedi2022interleaving, wang2023selfconsistency}.
However, to improve the reproducibility of our work we also provide results with an open-source LLM \cite{vicuna2023} and retriever \cite{khattab2020colbert}.

\section*{Acknowledgements}
We would like to thank Harsh Trivedi, Ofir Press, Mor Geva, Peter Clark and Ashish Sabharwal for their feedback and insightful comments. We thank SerpAPI for their support by granting us an academic discount.
This research was partially supported by the Yandex Initiative for Machine Learning and the European Research Council (ERC) under the European Union Horizons 2020 research and innovation programme (grants DELPHI 802800 and ProDIS 804302).
This work was completed in partial fulfillment of the Ph.D. of Ori Yoran and the Ph.D. of Tomer Wolfson.

\bibliography{anthology,custom}
\bibliographystyle{acl_natbib}

\appendix

\section{Evaluation}
\label{sec:evaluation_appendix}

\subsection{Generating Unknown as the Answer}
\label{sec:evaluation_abstain}
As we prompt LLMs to generate answers, a potential outcome is for the model to \emph{abstain} from answering the question, by generating \emph{Unknown} as its answer. Additional cases are when the model generates an end-of-sequence token without any final answer.
In the binary-choice datasets, \strategyqa{}, \quartz{} and \feverous{}, we assign a score of 0.5 to such examples, thereby simulating a random guess. 
When submitting predictions to the \strategyqa{} test set, we identify cases of model abstains or null predictions beforehand. For these examples, we assign a label of either \textit{Yes} or \textit{No} at random.
In datasets with open-ended answers, we assign a score of 0 when the predicted answer is either \emph{Unknown} or null.
To make Self-Ask a stronger baseline, when the greedy decoded chain has a null answer, we randomly choose a prediction from one of the other chains. For \selfconsistency{}, we do not consider predictions from chains where answers are \emph{Unknown} or null.

\subsection{\fermi{}}
\label{sec:fermi_appendix}
The \fermi{} dataset requires approximating numeric answers for open-ended questions. Example questions are shown in Tab.~\ref{tab:datasets} and Fig.~\ref{fig:architecture}.  When providing a \fermi{} question to our models and baselines we also add the gold answers measure units (e.g. meters, cubes, litres, etc.). While this additional input helps the model, we note that we provide it to all our baselines for a fair comparison with \mte{}. 
Nevertheless, even when given the gold units, predicting the final answers to \fermi{} problems remains highly challenging.

\section{Models}
\label{sec:appendix_models}

\subsection{Retrieval}
\label{sec:retrieval}
\begin{figure}[t]\setlength{\belowcaptionskip}{-8pt}
  \centering
  \includegraphics[width=0.49\textwidth]{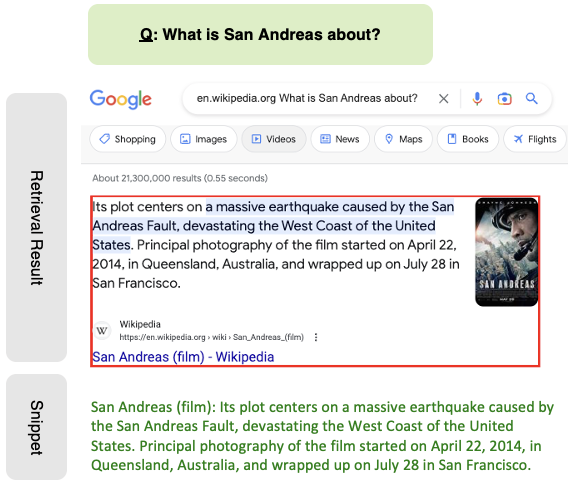}
  \caption{Example for a retrieved evidence snippet for one of the intermediate questions from Fig.~\ref{fig:multi_chain_example}.}
  \label{fig:snippet_example}
\end{figure}

For our retrieval, we use the Google Search Engine, via SerpAPI, and return the top-1 retrieved result as an evidence snippet. Snippets can include answer-boxes and tables.\footnote{https://serpapi.com/organic-results} We prepend the page title to the beginning of the snippet, as shown in Fig.~\ref{fig:snippet_example}.

\begin{table*}[t]
\centering
\footnotesize
  \begin{tabular}{p{4.2cm}ccc|ccc}
    \toprule
    Model & Ret. & LLM & \# Chains & \textsc{StrgyQA} & \hotpotqa & \wikihop{}\\
    \midrule
    CoT \cite{wang2023selfconsistency} & no & \codex{} & 1 & 73.4 & 39.8 & - \\
    CoT$+$SC$@40$ \cite{wang2023selfconsistency}& no & \codex{} & 40 & 79.8 & 44.6 & - \\
    Self-Ask \cite{press2023measuring} & yes & \davinci{} & 1 & - & - & 52.6 \\
    DSP \cite{Khattab2022DemonstrateSearchPredictCR} & yes & \davinci{} & 20 & - & 62.9 & -\\
    IR-CoT \cite{trivedi2022interleaving} & yes & \codex{} & 1 & - & 61.2 & 65.2 \\
    \midrule
    Self-Ask (ours) & yes & \codex{} & 1 & 69.3 & 50.2 & 63.8 \\

    \mte{} & yes & \codex{} & 5 & 73.6 (+4.3) & 57.0 (+6.8) & 67.9 (+4.1)\\
    \mte$+$\selfconsistency{}$@3$ & yes & \codex{} & 15 & 76.4 (+7.1) & 59.2 (+9.0) & 68.6 (+4.8) \\

  \bottomrule
\end{tabular}
\caption{Recent ODQA results using CoT prompting on LLMs. We list whether a retrieval component is used (Ret.), the LLM, and the number of reasoning chains used to generate the answer. Different systems evaluate on different evaluation sets for each dataset. Retrieval-augmented systems vary in terms of their retriever and corpus.}
  \label{tab:past_work}
\end{table*}

\subsection{Implementation Details}
\label{sec:hps}

We describe the design choices made in our \mte{} model, such as preforming retrieval on the original question and a variant of the meta-reasoner prompt for \feverous{}. Due to cost limitations, we evaluate our design choices at a smaller scale and avoid running an exhaustive grid search.

\paragraph{Retrieving the Original Question}
We follow past work \cite{trivedi2022interleaving} by incorporating retrieved evidence for the original question in addition to evidence retrieved for the intermediate steps (\S\ref{sec:method_decomp_retrieve}). This has a positive or negligible effect on most datasets however, it dramatically decreases the results of all models on the \fermi{} task. Results
drop for \selfask{} (38.3$\pm$0.7 to 34.7$\pm$0.5), \selfconsistency{} (38.3$\pm$0.8 to 34.4$\pm$0.3), \ste{} (38.1$\pm$0.8 to 34.4$\pm$0.8) and \mte{} (38.9$\pm$0.8 to 37.0$\pm$0.7).
Therefore, our models are run without original question retrieval when evaluated on \fermi{}.
Interestingly, while all models perform roughly the same without original question retrieval, \mte{} appears better by 2 points when evidence for the original question is used. We hypothesize that it might be due to \mte{} being somewhat more robust to the addition of irrelevant evidence. 

\paragraph{\feverous{} Meta-Reasoner Prompt}
As described in \S\ref{sec:method_entail}, the meta-reasoner generates an explanation which precedes the final answer. \feverous{} is distinct from all other datasets as it require verification of multiple facts in order to verify or disprove a complex statement. When a statement is false, we list one or more of its false intermediate facts along with its correction. For example, in Fig.~\ref{fig:entailment_prompt} we list that Robert Broderip lived in Bristol, not London. When prompting the meta-reasoner to list both true and false intermediate facts, we observed a decrease in performance for both \mte{} (69.4$\pm$1.0 to 66.4$\pm$0.7) and \ste{} (65.1$\pm$0.4 to 62.9$\pm$0.3). We hypothesize that repeating multiple true facts excessively prompts the model to predict the label ``Yes'' in cases where most (but not all) of the intermediate facts are correct.


\subsection{Empirical Comparison to Recent Approaches}
\label{sec:recent_approaches_comparison}

In Tab.~\ref{tab:past_work}, we compare \mte{} to recent CoT-based approaches for multi-hop reasoning. 
An apples-to-apples comparison is not possible, as these methods do not evaluate on all \numdatasets{} of our datasets and use varying samples of 500-1,000 dev examples for their evaluation. 
Moreover, different methods use different retrieval corpora, hyperparameters, prompts and LLMs. 
Nevertheless, we argue that a direct comparison serves as a measuring stick for \mte{}'s robustness across multiple datasets, compared to similar solutions.

Evaluation differences include the retrieval corpora, as both IR-CoT and DSP use the official Wikipedia dump provided with the \hotpotqa{} dataset \cite{yang2018hotpotqa}. Our retrieved evidence are from an updated version of Wikipedia, via Google Search. Since certain facts may change over time, this could potentially explain the high percentage of \mte{} predictions labeled as valid in our error analysis (\S\ref{sec:analysis}).

We emphasize that our focus is on highlighting the potential of reasoning on reasoning chains. \mte{} is a method aimed at improving models which generate reasoning chains. Compared to \selfconsistency{}, we observe that \mte{} further boosts the underlying \selfask{} model.
While task-specific improvements are possible, they are orthogonal to our work.

\subsection{Reasoning on Retrieved Evidence}
\label{sec:ent_contexts}
The meta-reasoner answers questions given a multi-chain context of \emph{question-answer} ($q_i$, $a_i$) pairs, extracted from multiple reasoning chains (\S\ref{sec:method_entail}). We experiment with an alternative multi-chain context, comprised of \emph{questions} and retrieved \emph{evidence} ($q_i$, $e_i$) (\S\ref{sec:method_decomp_retrieve}). 
This setting resembles past work \cite{trivedi2022interleaving} however, our sentences are intermediate evidence from \emph{multiple} reasoning chains, not just the greedy-decoded chain. We compare these variants, \mtec{} and \stec{}, to \mte{} and \ste{} that reason on QA pairs.
Tab.~\ref{tab:entailment_on_contexts} shows that meta-reasoning on retrieved evidence is less effective. 
The gap is more evident in implicit reasoning tasks, perhaps due to retrieved evidence being less relevant on average. Example prompts for \mtec{} and \stec{} are listed in \S\ref{sec:appendix_prompts}.

\begin{table*}[t]
\centering
\footnotesize
  \begin{tabular}{c|cc|cc}
    \toprule
    Dataset  & \stec  & \ste & \mtec  & \mte \\
    \midrule

    \strategyqa & 69.1$\pm$0.4 & \bf 70.0$\pm$0.6 & 73.2$\pm$0.6 & \bf73.6$\pm$0.7 \\
    \fermi & 34.1$\pm$0.6 & \bf38.1$\pm$0.8 & 33.9$\pm$0.3 & \bf38.9$\pm$0.8 \\
    \quartz  & 76.1$\pm$0.2 & \bf80.7$\pm$0.1 & 76.2$\pm$1.9 & \bf81.6$\pm$1.3 \\
    \midrule
    \hotpotqa & 53.5$\pm$0.1 & \bf 56.4$\pm$0.4 & \bf 58.2$\pm$1.0 & 57.0$\pm$0.8\\
    \wikihop & 66.2$\pm$0.2 & \bf67.2$\pm$0.2 & 67.1$\pm$0.9  & \bf67.9$\pm$0.4\\
    \bamboogle & 64.1$\pm$0.0 & \bf64.7$\pm$0.4 & \bf67.4$\pm$2.3 & 66.5$\pm$1.7\\
    \feverous & 64.0$\pm$0.5 & \bf 65.1$\pm$0.4 & 62.5$\pm$0.5 & \bf 69.4$\pm$1.0\\
  \bottomrule
\end{tabular}
\caption{Effect of using question-answer pairs versus question-evidence pairs as input to the \entailment{}.
}
  \label{tab:entailment_on_contexts}
\end{table*}

\section{Analysis}
\label{sec:appendix_analysis}

\subsection{When are Multiple Chains Helpful?}
\label{sec:appendix_multiple_chains}
\begin{figure*}[t]\setlength{\belowcaptionskip}{-8pt}
  \centering
  \includegraphics[trim={3.5cm 3.5cm 3.5cm 3cm}, clip, width=0.99\textwidth]{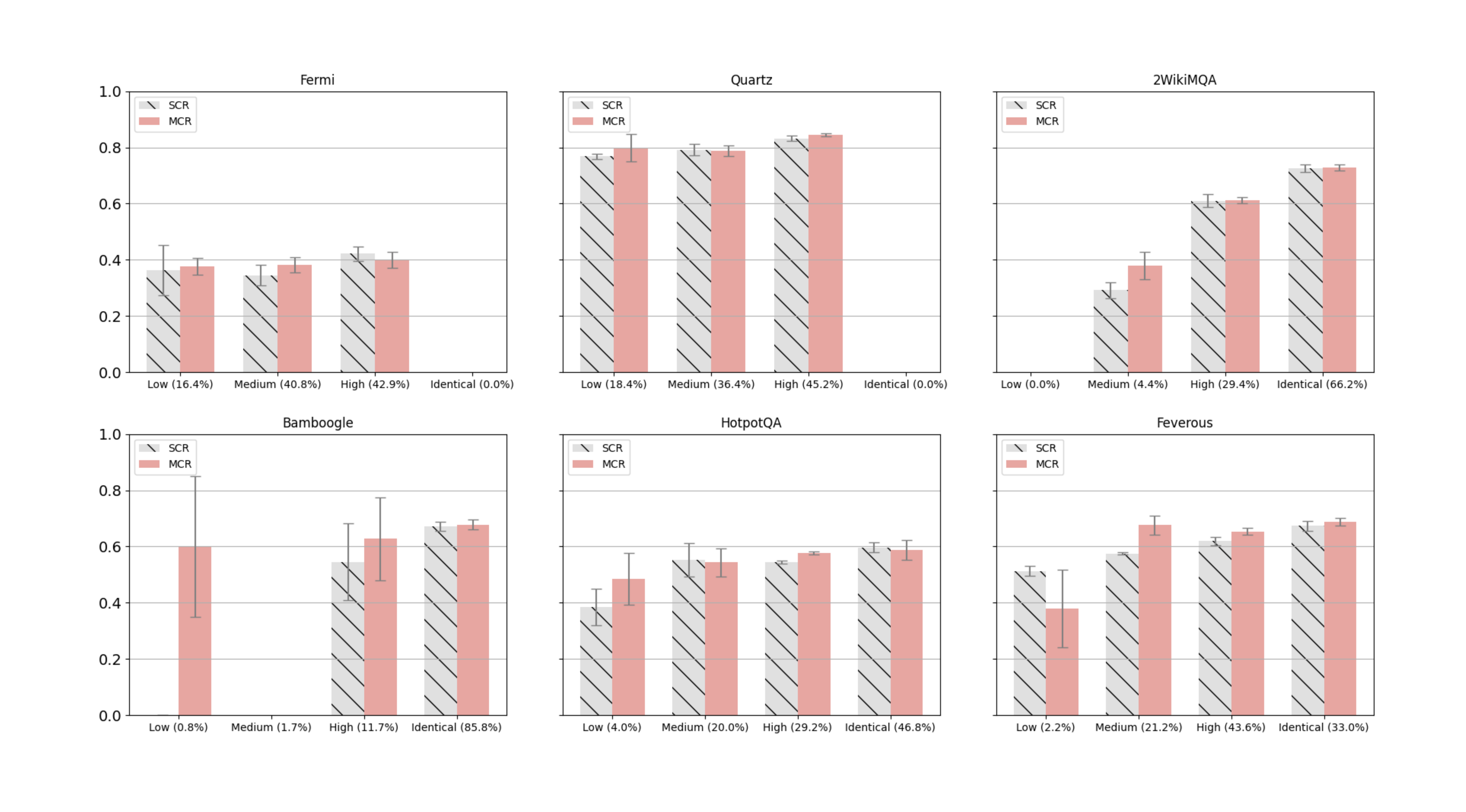}
  \caption{\mte{} and \ste{} accuracy on \fermi{}, \quartz{}, \wikihop{}, \bamboogle{}, \hotpotqa{}, and \feverous{}, on examples categorized by their \mte{} explanation's similarity to the greedy chain. \mte{} performs similarly to \ste{} when similarity is high, and outperfoms \ste{} when similarity is lower. Error bars indicate standard deviation, which tends to be high when the number of examples in the bin is small. For \feverous{} we display the variant where \mte{} has to repeat all relevant facts (\S\ref{sec:hps}), to make sure the \mte{} explanation is not empty.}
  \label{fig:mte_vs_ste_all_dataets}
\end{figure*}

In \S\ref{sec:analysis}, we have shown that the advantage of \mte{} over \ste{} lies in examples where the meta-reasoner uses chains other than the one generated through greedy decoding. In Fig.~\ref{fig:mte_vs_ste_all_dataets} we provide the results for all other datasets, in addition to the \strategyqa{} results in Fig.~\ref{fig:sim_to_greedy}. The similar trend among all datasets is that in examples with lower similarity to the greedy chain, \mte{} gains over \ste{} are higher. 

The similarity between the meta-reasoner explanation and the greedy decoded reasoning chain is defined as follows: We calculate the ROUGE-1-precision \cite{lin-2004-rouge} between the explanation and the chain. Low, Medium, and High are based on thresholds of $\frac{1}{3}$, $\frac{2}{3}$, and $1$ respectively, with the Identical category indicating an exact match.

\subsection{Combining Reasoning Chains}
\label{sec:combining_chains}
\begin{figure*}[t]\setlength{\belowcaptionskip}{-8pt}
  \centering
  \includegraphics[width=0.99\textwidth]{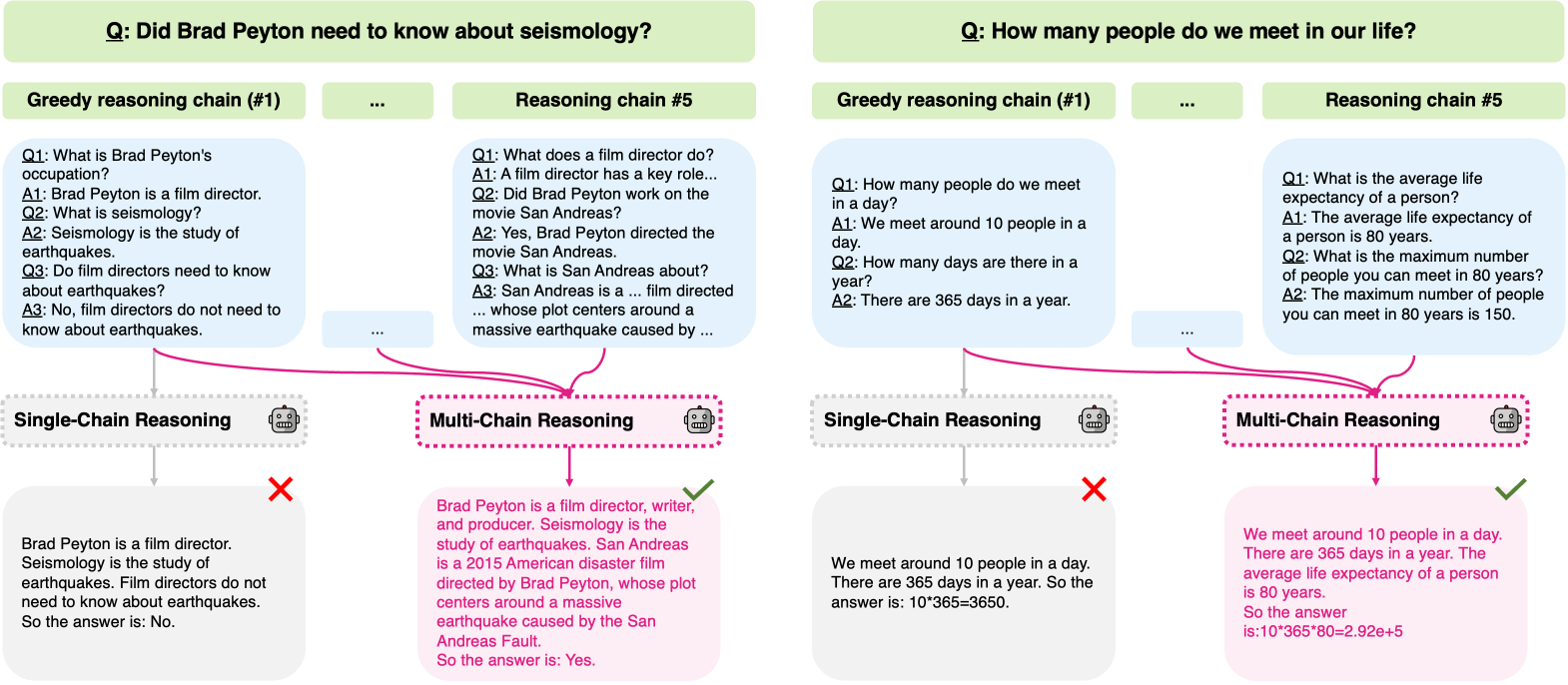}
  \caption{Examples for combining facts from multiple reasoning chains.}
  \label{fig:mixing_traces_examples}
\end{figure*}

Fig.~\ref{fig:mixing_traces_examples} provides additional examples for combining facts between multiple reasoning chains. 

\subsection{Explanation Quality Analysis}
\label{sec:explanation_quality_analysis}

\begin{figure}[t]\setlength{\belowcaptionskip}{-8pt}
  \centering
  \includegraphics[width=0.48\textwidth]{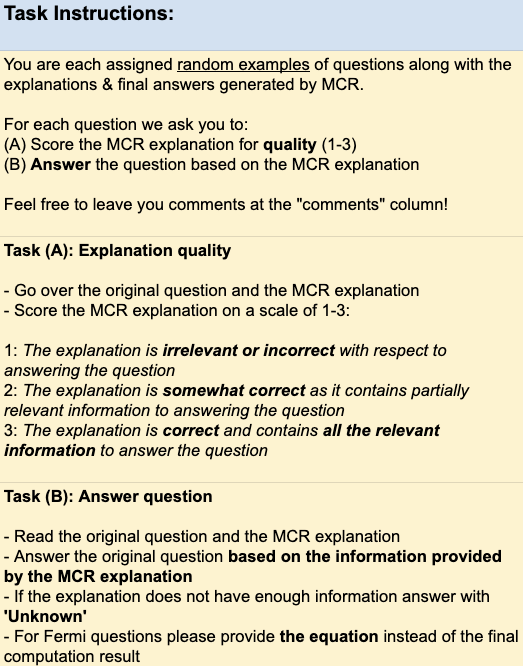}
  \vspace*{-0.7cm}
  \caption{The annotation instructions for the \mte{} explanation quality analysis.}
  \label{fig:ex_quality_instructions}
\end{figure}

We provide additional details on the annotation for the scoring meta-reasoner explanations. The annotation was performed by 4 graduate students that are authors of this paper. The annotators were presented with a question and an explanation, and asked to perform two tasks: (a) score the explanation for its quality and (b) answer the question based on the meta-reasoner explanation. We provide the full instructions shown to the annotators in Fig.~\ref{fig:ex_quality_instructions} and the full results in Tab.~\ref{tab:quality_analysis}.


\begin{table*}[t]
\centering
\footnotesize
  \begin{tabular}{ll|ccc|c|cc}
    \toprule
\textbf{dataset}  & Reasoning & \textbf{\%3} & \textbf{\%2} & \textbf{\%1} & \textbf{Sim\_predictions} & \textbf{Human\_acc} & \textbf{\mte{}\_acc} \\
\midrule
\strategyqa{} & & 79 & 18 & 3 & 89.9 & 77.2 & 72.2 \\
\fermi{} & implicit & 60 & 32 & 8 & 76.0 & 47.8 & 40.7 \\
\quartz{} & & 91 & 7 & 2 & 98.9 & 87.9 & 86.8 \\
\midrule
\hotpotqa{} & & 77 & 21 & 2 & 95.4 & 49.2 & 49.2 \\
\wikihop{} & explicit & 97 & 2 & 1 & 94.9 & 69.6 & 69.2 \\
\bamboogle{} & & 90 & 9 & 1 & 98.2 & 71.5 & 71.4 \\
\midrule
Average & implicit & 76.7 & 19.0 & 4.3 & 88.3 & 71.0 & 66.6 \\
Average & explicit & 88.0 & 10.7 & 1.3 & 96.2 & 63.4 & 63.3 \\
Average &&  82.3 & 14.8 & 2.8 & 92.2 & 67.2 & 64.9 \\
 \bottomrule
\end{tabular}
\caption{Full results  for the explanation quality analysis. \emph{Sim\_predictions} indicates the similarity between the human and \mte{} prediction, calculated using the dataset-specific metrics described in \S\ref{sec:datasets}. 
\emph{Human\_acc} and \emph{\mte{}\_acc} represent the accuracy of humans and \mte{} predictions, respectively. Since only explanations with a score of 3 are guaranteed to contain the necessary information to arrive at an answer, we filter other examples when calculating \emph{sim\_predictions}, \emph{Human\_acc}, and \emph{\mte{}\_acc}.
}
  \label{tab:quality_analysis}
\end{table*}

\subsection{Error Analysis}
\label{sec:error_analysis}
We provide additional details regarding our error analysis (\S\ref{sec:analysis}). 
In less than 5\%, we encountered grammatically bad questions which we were unable to comprehend and were therefor discarded from our analysis. For example the \hotpotqa{} question: ``What does the goddess associated with the goddess Frigg consists of what tales?''

The input to our \entailment{} model is a context comprised of $(q_i, a_i)$ pairs, generated by the \decomposition{} model. As the \decomposition{} model is an LLM that is conditioned on retrieved evidence (and prior decomposition steps) it may hallucinate false intermediate answers. In cases of such hallucinations we distinguish between two error types, based on the relevant component. First, \emph{Retrieval errors} are cases where no relevant information was retrieved, leading to the \decomposition{} model hallucinating an incorrect $a_i$,  passed on to the meta-reasoner's context. Second, we treat cases where relevant evidence was retrieved, but the \decomposition{} model ignored it and hallucinated an incorrect $a_i$ as \emph{Decomposition errors}. 

Errors stemming from \emph{Contradicting Facts}, are cases where the meta-reasoner context contains two contradicting facts, one accurate while the other was hallucinated by the \decomposition{} model. For example, Fig.~\ref{fig:contradicting_facts} displays an example where the context has contradicting facts on who was the father of Eliezer Ben-Yehuda. When the meta-reasoner has contradicting facts, it is expected to select the correct fact, based on the knowledge encoded in its parameters. Addressing such errors in future work could rely on refining generated text with methods such as RARR \cite{gao2022rarr}. 

As our error classes mainly match the \mte{} components, this error breakdown could potentially help to guide future improvements.

\begin{figure}[t]\setlength{\belowcaptionskip}{-8pt}
\footnotesize
  \begin{tabular}{p{0.97\linewidth}}
    \toprule \bf
    \textit{Given a question and a context, answer the question step-by-step. If you are unsure, answer Unknown.}\\\midrule
    \bf Context:\\
    Who is the father of modern Hebrew? The father of modern Hebrew is Eliezer Ben-Yehuda.\\
    Who is the father of Eliezer Ben-Yehuda? The father of Eliezer Ben-Yehuda is Abraham.\\
    ...\\
    Who is the father of modern Hebrew? The father of modern Hebrew is Eliezer Ben-Yehuda.\\
    Who is the father of Eliezer Ben-Yehuda? Eliezer Ben-Yehuda's father is Yehuda Leib.
    \\\midrule
    \textbf{Question:} Who is the father of the father of modern Hebrew?
 \\\midrule
    \textbf{Answer:} The father of modern Hebrew is Eliezer Ben-Yehuda. The father of Eliezer Ben-Yehuda is Abraham. \textbf{So the answer is:} Abraham.\\
    \midrule
    \textbf{Gold answer is:} \textit{Yehuda Leib} \\
     \bottomrule
\end{tabular}
\caption{Example a \emph{Contradicting Facts} error. When generating the explanation, the meta-reasoner has to rely on knowledge encoded in its parameters to decide between multiple contradicting facts in its context on who was the father of Eliezer Ben-Yehuda.}
\label{fig:contradicting_facts}
\end{figure}

\section{Prompts}
\label{sec:appendix_prompts}

\subsection{Prompt Details}
\label{sec:appendix_prompts_details}

We provide example prompts for our models for one explicit dataset (\wikihop{}, \decomposition: Fig.~\ref{fig:wikihop_decomposition_prompt}, \mte{}/\ste{}: Fig.~\ref{fig:wikihop_entailment_prompt}, \mtec{}/\stec{}: Fig.~\ref{fig:wikihop_entailment_c_prompt}) and one implicit dataset (\strategyqa{}, \decomposition: Fig.~\ref{fig:strategyqa_decomposition_prompt}, \mte{}/\ste{}: Fig.~\ref{fig:strategyqa_entailment_prompt}, \mtec{}/\stec{}:Fig.~\ref{fig:strategyqa_entailment_c_prompt}). 
All of our prompts will be released along with our codebase. We use random examples and spend minimal effort on prompt engineering. The number of exemplars varies slightly between dataset and model, with the exact numbers listed in Tab.~\ref{tab:num_examplars}.

\begin{table*}[t]
\centering
\footnotesize
  \begin{tabular}{ccccccc}
    \toprule
    Dataset & \selfask{} & \ste & \mte & \stec & \mtec  \\
    \midrule

    \strategyqa & 10 & 6 & 6 & 6 & 6  \\
    \fermi  & 6 & 6 & 6 & 6 & 6  \\
    \quartz  & 6 & 8 & 8 & 8 & 8  \\
    \hotpotqa & 12 & 10 & 10 & 10 & 10  \\
    \wikihop & 6 & 6 & 6 & 6 & 6  \\
    \bamboogle & 6 & 6 & 6 & 6 & 6  \\
    \feverous & 10 & 10 & 10 & 10 & 10  \\
  \bottomrule
\end{tabular}
\caption{The number of exemplars for each model and dataset. Since \mte{} and \ste{}, and \mtec{} and \stec{} use the same prompt they have the same number of exemplars. 
}
  \label{tab:num_examplars}
\end{table*}

\subsection{Prompt Statistics}
\label{sec:appendix_prompts_stats}
In Tab.~\ref{tab:prompt_len_stats} we provide statistics of the sequence lengths for all of our models, which include all the decomposition prompts, output decomposition sequences, retrieved evidence and the meta-reasoning prompts. The statistics are for our decomposition model (used by all of our baselines), as well as for the meta-reasoning prompts (used by \ste{} and \mte{}). Note that generating a single reasoning chain requires multiple LLM calls, one for each decomposition step. Therefore, a single decomposition generation is generally longer than applying one additional meta-reasoning step. 

Results are averaged over multiple runs, corresponding to the results in Tab.~\ref{tab:dev_set_results}.  Sequence lengths in Tab.~\ref{tab:prompt_len_stats} correspond to the number of tokens provided by the \codex{} tokenizer. 

\begin{table*}[t]
\centering
\footnotesize
  \begin{tabular}{cccccccc}
    \toprule
    Dataset & Dec. & Dec. steps & Dec. out. & Ret. len. & Meta-reason & \ste{} & \mte{} \\
    \midrule

    \strategyqa & 2,242 & 2.9±0.6	 & 103.3± 48.0 & 190.6±66.7 & 1,652 & 1,749.0± 52.0 & 2,032.9±110.7 \\
    \fermi & 1,442 & 2.3±0.9 & 91.4±31.9 & 165.8±78.9 & 1,681 & 1,765.5±25.5 & 1,984.1±79.6 \\
    \quartz &  839 & 1.2±0.5 & 55.2±20.6 & 92.7±28.7 & 2,129 & 2,202.4±19.8 & 2,343.6±48.3\\
    \hotpotqa &  2,508 & 1.7±0.7 & 86.1±91.9 & 153.0±84.1 & 2,380 & 2,460.3±28.1 & 2,666.2±90.0 \\
    \wikihop & 1,920 & 2.4±0.8 & 92.7±30.5 & 201.3±59.3 & 2,029 & 2,116.4±25.6 & 2,363.0±104.6 \\
    \bamboogle & 1,342 & 2.0±0.3 & 74.3±37.5 & 204.5±72.1 & 966 & 1,035.1±12.9 & 1,223.1±50.2 \\
    \feverous & 3,741 & 2.9±0.9 & 118.2±36.4 & 197.1±69.2 & 2,826 & 2,956.8±38.1 & 3,276.8±123.1	\\\midrule
    Average & 2,004.9 & 2.2±0.6 & 88.7±18.7 & 172.1±37.0 & 1,951.9 & 2,040.8±563.1 & 2,270.0±587.7 \\
  \bottomrule
\end{tabular}
\caption{Prompt lengths (number of tokens) used for each dataset: decomposition prompts (Dec.); number of output decomposition steps (Dec. steps); output decomposition length (Dec. out.); retrieved evidence length (Ret. len.); meta-reasoning prompt; \ste{} prompt length; \mte{} prompt length.}
  \label{tab:prompt_len_stats}
\end{table*}

\subsection{Robustness to Choice of Prompt}
\label{sec:appendix_prompts_sensitivity}
We empirically measure our method's sensitivity to the prompt of choice. To this end, we randomly sampled new exemplars for both our decomposition and meta-reasoning prompts for \strategyqa{} and \hotpotqa{}. When using different random exemplars, we observe that \mte{} still outperforms all baselines. Even though decomposition performance (\selfask{}) is more affected by the set of exemplars, the performance trend remains the same, with \mte{} being on top. Tab.~\ref{tab:prompt_sensitivity} lists the experiment results, evaluated on 500 examples from each dataset. We also provide the original prompt results in parenthesis (averaged over 3 runs).

\begin{table*}[t]
\centering
\footnotesize
  \begin{tabular}{ccccccccc}
    \toprule
    Dataset & Examples & \selfask{} & \selfconsistency{}$@5$  & \ste & \mte \\
    \midrule

    \strategyqa & 500 & 66.6 
 (69.3$\pm$0.3) & 71.0 (72.2$\pm$0.8) & 68.2 (70.0$\pm$0.6) & \textbf{72.0 (73.6$\pm$0.7)}\\
    \hotpotqa & 500 & 54.3 (50.2$\pm$0.3) & 56.2 (51.3$\pm$0.2) & 57.1 (56.4$\pm$0.4) & \textbf{59.3 (57.0$\pm$0.8)}\\
  \bottomrule
\end{tabular}
\caption{Experiments with \codex{} when using prompts with different random exemplars for both the decomposition and meta-reasoning prompts. Original prompt results are in parenthesis, for comparison. }
  \label{tab:prompt_sensitivity}
\end{table*}

\begin{figure*}[t]\setlength{\belowcaptionskip}{-8pt}
\footnotesize
  \begin{tabular}{p{0.97\linewidth}}
    \toprule \bf
    \textit{Given the following question, answer it by providing follow up questions and intermediate answers. If no follow up questions are necessary, answer the question directly. You are also provided with the most relevant google snippet for each intermediate question.}\\\midrule
\#\\
Context1: Xawery Żuławski: Polish-Russian War (Wojna polsko-ruska) is a 2009 Polish film directed by Xawery Żuławski based on the novel Polish-Russian War under the white-red flag by Dorota Masłowska. So the answer is Xawery Żuławski.\\
Context2: Xawery Żuławski: Xawery Żuławski ; National Film School in Łódź · 1995–present · Maria Strzelecka · 2.\\
Question: Who is the mother of the director of film Polish-Russian War (Film)?\\
Are follow up questions needed here: Yes.\\
Follow up: Who is the director of the film Polish-Russian War (Film)?\\
Intermediate answer: The director of the film Polish-Russian War is Xawery Żuławski.\\
Follow up: Who is the mother of Xawery Żuławski?\\
Intermediate answer: The mother of Xawery Żuławski is Małgorzata Braunek.\\
So the final answer is: Rick Scott Małgorzata Braunek.\\
... \\
\#\\
Context1: John, Count Palatine of Neumarkt: John (Johann von Pfalz-Neumarkt; 1383 – 14 March 1443) was the Count Palatine of Neumarkt from 1410 to his death. The son of Rupert III of the Palatinate, he married Catherine of Pomerania in 1407.\\
Context2: John, Count Palatine of Neumarkt: John (Johann von Pfalz-Neumarkt; 1383 – 14 March 1443) was the Count Palatine of Neumarkt from 1410 to his death. The son of Rupert III of the Palatinate, he married Catherine of Pomerania in 1407.\\
Question: Who is Catherine Of Pomerania, Countess Palatine Of Neumarkt's father-in-law?\\
Are follow up questions needed here: Yes.\\
Follow up: Who is the husband of Catherine of Pomerania, Countess Palatine of Neumarkt?\\
Intermediate answer: The husband of Catherine of Pomerania, Countess Palatine of Neumarkt is John, Count Palatine of Neumarkt.\\
Follow up: Who is the father of John, Count Palatine of Neumarkt?\\
Intermediate answer: The father of John, Count Palatine of Neumarkt is Rupert III of the Palatinate.\\
So the final answer is: Rupert III of the Palatinate.\\
\#\\
Context1: Crimen a las tres: Crimen a las tres is a 1935 Argentine crime film directed and written by Luis Saslavsky. Crimen a las tres. Directed by, Luis Saslavsky.\\
Context2: Elio Petri: The Working Class Goes to Heaven (Italian: La classe operaia va in paradiso), released in the US as Lulu the Tool, is a 1971 political drama film directed by Elio Petri. So the answer is Elio Petri.\\
Context3: March 20, 1995: Luis Saslavsky (April 21, 1903 – March 20, 1995) was an Argentine film director, screenwriter and film producer, and one of the influential directors in the Cinema of Argentina of the classic era. So the answer is March 20, 1995.\\
Context4: Elio Petri: Final years. In 1981, Petri visited Geneva to direct Arthur Miller\'s new play The American Clock, with Marcello Mastroianni playing the lead role. Petri died of cancer on 10 November 1982. He was 53 years old.\\
Question: Which film has the director died first, Crimen A Las Tres or The Working Class Goes To Heaven?\\
Are follow up questions needed here: Yes.\\
Follow up: Who is the director of Crimen a las tres?\\
Intermediate answer: The director of Crimen a las tres is Luis Saslavsky.\\
Follow up: Who is the director of The Working Class Goes to Heaven?\\
Intermediate answer: The director of The Working Class Goes to Heaven is Elio Petri.\\
Follow up: When did Luis Saslavsky die?\\
Intermediate answer: Luis Saslavsky died on March 20, 1995.\\
Follow up: When did Elio Petri die?\\
Intermediate answer: Elio Petri died on 10 November 1982.\\
So the final answer is: The Working Class Goes to Heaven. \\
\#\\
\bottomrule
\end{tabular}
\caption{Instruction and exemplars for the \wikihop{} \decomposition{} prompt.}
\label{fig:wikihop_decomposition_prompt}
\end{figure*}

\begin{figure*}[t]\setlength{\belowcaptionskip}{-8pt}
\footnotesize
  \begin{tabular}{p{0.97\linewidth}}
    \toprule \bf
    \textit{Given the following question, answer it by providing follow up questions and intermediate answers. For each follow up question, you are given a context which is the top returned google snippet for the question from Wikipedia. If no follow up questions are necessary, answer the question directly.}\\\midrule
\#\\
Context1: Frost: Frost is a thin layer of ice on a solid surface, which forms from water vapor in an above-freezing atmosphere coming in contact with a solid surface whose ...\\
Context2: Graduation: Graduation is the awarding of a diploma to a student by an educational institution. It may also refer to the ceremony that is associated with it.\\
Context3: Winter: Winter ; Astronomical season, 22 December – 21 March ; Meteorological season, 1 December – 28/29 February ; Solar (Celtic) season, 1 November – 31 January.\\
Question: Is it common to see frost during some college commencements?\\
Are follow up questions needed here: Yes.\\
Follow up: What seasons can you expect to see frost?\\
Intermediate answer: Frost is common during the winter.\\
Follow up: When is college commencement?\\
Intermediate answer: College commencement ceremonies often happen during the months of December, May, June.\\
Follow up: Do any of the months December, May, June occur during the Winter?\\
Intermediate answer: December is in the winter.\\
So the final answer is: Yes.\\
...\\
\#\\
Context1: Last rites: The last rites, also known as the Commendation of the Dying, are the last prayers and ministrations given to an individual of Christian faith, when possible, shortly before death. They may be administered to those awaiting execution, mortally injured, or terminally ill.\\
Context2: Richard Dawkins: Dawkins is an outspoken atheist and a supporter of various atheist, secular, and humanist organisations, including Humanists UK and the Brights movement. Dawkins suggests that atheists should be proud, not apologetic, stressing that atheism is evidence of a healthy, independent mind.\\
Context3: Prayer in the Catholic Church: In the Catholic Church, prayer is "the raising of one\'s mind and heart to God or the requesting of good things from God." It is an act of the moral virtue ...\\
Question: Would Richard Dawkins hypothetically refuse an offering of the Last rites?\\
Are follow up questions needed here: Yes.\\
Follow up: What are the last Rites?\\
Intermediate answer: The Last rites, in Catholicism, are the last prayers and ministrations given to an individual of the faith, when possible, shortly before death.\\
Follow up: What are Richard Dawkins religious beliefs?\\
Intermediate answer:  Richard Dawkins is known as an outspoken atheist, well known for his criticism of creationism and intelligent design.\\
Follow up: Would an atheist participate in Catholics prayers?\\
Intermediate answer:  It is unlikely that an atheist would participate in Catholics prayers.\\
So the final answer is: Yes.\\
\#\\
Context1: number 1: Hydrogen is the chemical element with the symbol H and atomic number 1. Hydrogen is the lightest element. So the answer is number 1.\\
Context2: Spice Girls - Simple English Wikipedia, the free encyclopedia: The group has five members. Each member uses a nickname initially given to them: Melanie Chisholm ("Sporty Spice"), Emma Bunton ("Baby Spice"), Melanie Brown ("Scary Spice"), Victoria Beckham (née Adams) ("Posh Spice"), and Geri Halliwell ("Ginger Spice") .\\
Context3: Square number: In mathematics, a square number or perfect square is an integer that is the square of an integer; in other words, it is the product of some integer with ...\\
Question: Hydrogen's atomic number squared exceeds number of Spice Girls?\\
Are follow up questions needed here: Yes.\\
Follow up: What is the atomic number of hydrogen?\\
Intermediate answer: Hydrogen has an atomic number of 1.\\
Follow up: How many people are in the Spice Girls band?\\
Intermediate answer: The Spice Girls has 5 members.\\
Follow up: Is the square of 1 greater than 5?\\
Intermediate answer: The square of 1 is 1 which is less than 5.\\
So the final answer is: No.\\
\#\\
\bottomrule
\end{tabular}
\caption{Instruction and exemplars for the \strategyqa{} \decomposition{} prompt.}
\label{fig:strategyqa_decomposition_prompt}
\end{figure*}

\begin{figure*}[t]\setlength{\belowcaptionskip}{-8pt}
\footnotesize
  \begin{tabular}{p{0.97\linewidth}}
    \toprule \bf
    \textit{Given a question and a context, answer the question and explain why. If you are unsure, answer Unknown.}\\\midrule
    \#\\
Context:\\
Who is the wife of Douglas Leiterman? The wife of Douglas Leiterman is Beryl Fox.\\
Where was Beryl Fox born? Beryl Fox was born in Winnipeg, Manitoba.\\
When and where was Mary born? Mary was born in c. 18 BC or September 8 (21), 16 BC Herodian Kingdom of Judea.\\
Where was Beryl Fox born? Beryl Fox was born in Winnipeg, Manitoba.\\
Who is the wife of Douglas Leiterman? The wife of Douglas Leiterman is Mary.\\
Who is the wife of Douglas Leiterman? The wife of Douglas Leiterman is Beryl Fox.\\
\\
Question:\\
Where was the wife of Douglas Leiterman born?\\
\\
Answer:\\
The wife of Douglas Leiterman is Beryl Fox. Beryl Fox was born in Winnipeg, Manitoba.\\
So the answer is: Winnipeg, Manitoba.\\
...\\
\#\\
Context:\\
Who is Beatrice of Aragon's father? The father of Beatrice of Aragon is King Ferdinand I of Naples.\\
Who is the father of Rupert III, Elector Palatine? The father of Rupert III, Elector Palatine is Rupert II, Elector Palatine.\\
Who is the husband of Catherine of Pomerania? The husband of Catherine of Pomerania is John II, Count Palatine of Neumarkt.\\
Who is Catherine Of Pomerania, Countess Palatine Of Neumarkt's husband? The husband of Catherine Of Pomerania, Countess Palatine Of Neumarkt is John I, Count Palatine of Neumarkt.\\
Who is the father of John II, Count of Holstein-Rendsburg? The father of John II, Count of Holstein-Rendsburg is Henry II, Count of Holstein-Rendsburg.\\
Who is Catherine Of Pomerania, Countess Palatine Of Neumarkt's husband? The husband of Catherine Of Pomerania, Countess Palatine Of Neumarkt is John II, Count of Holstein-Rendsburg.\\
Who is the father of John I, Count Palatine of Neumarkt? The father of John I, Count Palatine of Neumarkt is Rupert III, Elector Palatine.\\
Who are the parents of Rupert III, Elector Palatine? The parents of Rupert III, Elector Palatine are Rupert II, Elector Palatine and Beatrice of Aragon.\\
Who is the father of John II, Count Palatine of Neumarkt? The father of John II, Count Palatine of Neumarkt is Rupert III, Elector Palatine.\\
\\
Question:\\
Who is Catherine Of Pomerania, Countess Palatine Of Neumarkt's father-in-law?\\
\\
Answer:\\
The husband of Catherine Of Pomerania, Countess Palatine Of Neumarkt is John I, Count Palatine of Neumarkt. The father of John I, Count Palatine of Neumarkt is Rupert III, Elector Palatine.\\
So the answer is: Rupert III, Elector Palatine.\\
\#\\
Context:\\
When did Elio Petri die? Elio Petri died on 10 November 1982.\\
Who is the director of The Working Class Goes to Heaven? The director of The Working Class Goes to Heaven is Elio Petri.\\
Who is the director of Crimen A Las Tres? The director of Crimen A Las Tres is Luis Saslavsky.\\
Who is the director of Crimen A Las Tres? The director of Crimen A Las Tres is Luis Saslavsky.\\
When did Luis Saslavsky die? Luis Saslavsky died on March 20, 1995.\\
Who is the director of Crimen A Las Tres? The director of Crimen A Las Tres is Luis Saslavsky.\\
When did Elio Petri die? Elio Petri died on 10 November 1982.\\
When did Luis Saslavsky die? Luis Saslavsky died on March 20, 1995.\\
When did Luis Saslavsky die? Luis Saslavsky died on March 20, 1995.\\
When did Elio Petri die? Elio Petri died on 10 November 1982.\\
Who is the director of The Working Class Goes to Heaven? The director of The Working Class Goes to Heaven is Elio Petri.\\
Who is the director of The Working Class Goes to Heaven? The director of The Working Class Goes to Heaven is Elio Petri.\\
\\
Question:\\
Which film has the director died first, Crimen A Las Tres or The Working Class Goes To Heaven?\\
\\
Answer:\\
The director of Crimen A Las Tres is Luis Saslavsky. The director of The Working Class Goes to Heaven is Elio Petri. Luis Saslavsky died on March 20, 1995. Elio Petri died on 10 November 1982.\\
So the answer is: The Working Class Goes To Heaven.\\
\#\\
\bottomrule
\end{tabular}
\caption{Instruction and exemplars for the \wikihop{} \entailment{} prompt.}
\label{fig:wikihop_entailment_prompt}
\end{figure*}

\begin{figure*}[t]\setlength{\belowcaptionskip}{-8pt}
\footnotesize
  \begin{tabular}{p{0.97\linewidth}}
    \toprule \bf
    \textit{Given a question and a context, provide a Yes or No answer and explain why. If you are unsure, answer Unknown.}\\\midrule
\#\\
Context:\\
What type of animal is a jellyfish? A jellyfish is a type of invertebrate.\\
Do jellyfish have arteries? No, jellyfish do not have arteries. They have a type of nervous system called a nerve net.\\
What is atherosclerosis? Atherosclerosis is a disease where plaque builds up in the arteries.\\
Do jellyfish have arteries? Jellyfish do not have a circulatory system and therefore do not have arteries.\\
So jellyfish don't have atherosclerosis because they don't have arteries? Yes, that is correct.\\
What is atherosclerosis? Atherosclerosis is a type of heart disease.\\
Is an invertebrate susceptible to atherosclerosis? No, invertebrates are not susceptible to atherosclerosis.\\
What is atherosclerosis? Atherosclerosis is a disease in which plaque builds up on the walls of the arteries.\\
\\
Question:\\
Is a jellyfish safe from atherosclerosis?\\
 \\
Answer:\\
Jellyfish do not have a circulatory system and therefore do not have arteries. Atherosclerosis is a disease in which plaque builds up on the walls of the arteries.\\
So the answer is: Yes.\\
\#\\
...\\
Context:\\
What is tonsure? Tonsure is the practice of cutting or shaving the hair from the scalp, as a sign of religious devotion or humility.\\
What is Christopher Hitchens' religious affiliation? Christopher Hitchens was an atheist.\\
What is tonsure? Tonsure is a practice of cutting or shaving the hair from the scalp.\\
Would Christopher Hitchens be likely to do something as a sign of religious devotion? No.\\
Would Christopher Hitchens be religious? Christopher Hitchens was an atheist.\\
Is an atheist likely to engage in religious practices? No, an atheist is not likely to engage in religious practices.\\
What is tonsure? Tonsure is the practice of cutting or shaving the hair from the scalp, as a sign of religious devotion or humility.\\
What is Christopher Hitchens' religious affiliation? Christopher Hitchens was an atheist.\\
Do atheists usually engage in tonsure? No, atheists do not usually engage in tonsure as it is a religious practice.\\
 \\
Question:\\
Would Christopher Hitchens be very unlikely to engage in tonsure?\\
 \\
Answer:\\
Tonsure is the practice of cutting or shaving the hair from the scalp, as a sign of religious devotion or humility. Christopher Hitchens was an atheist. An atheist is not likely to engage in religious practices.\\
So the answer is: Yes.\\
\#\\
Context:\\
What is the evidence that Neanderthals used arithmetic? There is no direct evidence that Neanderthals used arithmetic, but indirect evidence suggests that they may have had some understanding of basic mathematical concepts.\\
What is the earliest evidence of arithmetic? The earliest evidence of arithmetic dates back to the ancient Sumerians, who lived in what is now modern-day Iraq.\\
Did Neanderthals live before the ancient Sumerians? Yes.\\
What is arithmetic? Arithmetic is the branch of mathematics that deals with the properties and manipulation of numbers.\\
Is there any reason to believe that Neanderthals could not have used arithmetic? There is no reason to believe that Neanderthals could not have used arithmetic.\\
Do we have any evidence that Neanderthals could manipulate numbers? We have no evidence that Neanderthals could manipulate numbers.\\
What is arithmetic? Arithmetic is the branch of mathematics that deals with the properties and manipulation of numbers.\\
What is arithmetic? Arithmetic is the branch of mathematics that deals with the addition, subtraction, multiplication, and division of whole numbers and with the manipulation of fractions and decimals.\\
 \\
Question:\\
Did Neanderthals use arithmetic?\\
 \\
Answer:\\
There is no evidence that Neanderthals used arithmetic. The earliest evidence of arithmetic dates back to the ancient Sumerians who lived after the Neanderthals.\\
So the answer is: No.\\
\#\\
\bottomrule
\end{tabular}
\caption{Instruction and exemplars for the \strategyqa{} \entailment{} prompt.}
\label{fig:strategyqa_entailment_prompt}
\end{figure*}

\begin{figure*}[t]\setlength{\belowcaptionskip}{-8pt}
\footnotesize
  \begin{tabular}{p{0.97\linewidth}}
    \toprule \bf
    \textit{Given a question and a context, answer the question and explain why. If you are unsure, answer Unknown.}\\\midrule
    \#\\
Context:\\
Who is the wife of Douglas Leiterman? | Douglas Leiterman (1927 – 19 December 2012) was a Canadian television producer.: [[\'Douglas Leiterman\'], [\'Spouse\', \'Beryl Fox\']]\\

Where was Beryl Fox born? | Winnipeg, Manitoba: Fox was born in 1931 in Winnipeg, Manitoba. So the answer is Winnipeg, Manitoba.\\

Who is Douglas Leiterman's wife? | Douglas Leiterman (1927 – 19 December 2012) was a Canadian television producer.: [[\'Douglas Leiterman\'], [\'Known for\', \'This Hour Has Seven Days\'], [\'Spouse\', \'Beryl Fox\']]\\

Where was Beryl Fox born? | Winnipeg, Manitoba: Fox was born in 1931 in Winnipeg, Manitoba. So the answer is Winnipeg, Manitoba.\\

Who is the wife of Douglas Leiterman? | Douglas Leiterman (1927 – 19 December 2012) was a Canadian television producer.: [[\'Douglas Leiterman\'], [\'Spouse\', \'Beryl Fox\']]\\

When and where was Mary born? | Mary was born in c. 18 BC or September 8 (21), 16 BC Herodian Kingdom of Judea.\\
\\
Question:\\
Where was the wife of Douglas Leiterman born?\\
\\
Answer:\\
The wife of Douglas Leiterman is Beryl Fox. Beryl Fox was born in Winnipeg, Manitoba.\\
So the answer is: Winnipeg, Manitoba.\\
...\\
\#\\
Context:\\
Who is the director of Crimen A Las Tres? | Crimen a las tres: Crimen a las tres is a 1935 Argentine crime film directed and written by Luis Saslavsky.\\

Who is the director of The Working Class Goes to Heaven? | The Working Class Goes to Heaven: The Working Class Goes to Heaven (Italian: La classe operaia va in paradiso), released in the US as Lulu the Tool, is a 1971 political drama film directed by Elio Petri.\\

When did Luis Saslavsky die? | Luis Saslavsky: Luis Saslavsky (April 21, 1903 – March 20, 1995) was an Argentine film director, screenwriter and film producer, and one of the influential directors in the Cinema of Argentina of the classic era.\\

When did Elio Petri die? | Elio Petri: Petri died of cancer on 10 November 1982. He was 53 years old.\\

Who is the director of Crimen A Las Tres? | Crimen a las tres: Crimen a las tres is a 1935 Argentine crime film directed and written by Luis Saslavsky.\\

Who is the director of The Working Class Goes to Heaven? | The Working Class Goes to Heaven: The Working Class Goes to Heaven (Italian: La classe operaia va in paradiso), released in the US as Lulu the Tool, is a 1971 political drama film directed by Elio Petri.\\

When did Luis Saslavsky die? | Luis Saslavsky: Luis Saslavsky (April 21, 1903 – March 20, 1995) was an Argentine film director, screenwriter and film producer, and one of the influential directors in the Cinema of Argentina of the classic era.\\

When did Elio Petri die? | Elio Petri: Petri died of cancer on 10 November 1982. He was 53 years old.\\

Who is the director of Crimen A Las Tres? | Crimen a las tres: Crimen a las tres is a 1935 Argentine crime film directed and written by Luis Saslavsky.\\

When did Luis Saslavsky die? | Luis Saslavsky: Luis Saslavsky (April 21, 1903 – March 20, 1995) was an Argentine film director, screenwriter and film producer, and one of the influential directors in the Cinema of Argentina of the classic era.\\

Who is the director of The Working Class Goes to Heaven? | The Working Class Goes to Heaven: The Working Class Goes to Heaven (Italian: La classe operaia va in paradiso), released in the US as Lulu the Tool, is a 1971 political drama film directed by Elio Petri.\\

When did Elio Petri die? | Elio Petri: Petri died of cancer on 10 November 1982. He was 53 years old.\\
\\
Question:\\
Which film has the director died first, Crimen A Las Tres or The Working Class Goes To Heaven?\\
\\
Answer:\\
The director of Crimen A Las Tres is Luis Saslavsky. The director of The Working Class Goes to Heaven is Elio Petri. Luis Saslavsky died on March 20, 1995. Elio Petri died on 10 November 1982.\\
So the answer is: The Working Class Goes To Heaven.\\
\#\\
\bottomrule
\end{tabular}
\caption{Instruction and exemplars for the \wikihop{} \entailment{} prompt for \mtec{} and \stec{} reasoning over retrieved evidence.}
\label{fig:wikihop_entailment_c_prompt}
\end{figure*}

\begin{figure*}[t]\setlength{\belowcaptionskip}{-8pt}
\footnotesize
  \begin{tabular}{p{0.97\linewidth}}
    \toprule \bf
    \textit{Given a question and a context, answer the question step-by-step. If you are unsure, answer Unknown.}\\\midrule
    \#
\\
Context:\\
What is atherosclerosis? | Atherosclerosis: Atherosclerosis is a pattern of the disease arteriosclerosis in which the wall of the artery develops abnormalities, called lesions. These lesions may lead to narrowing due to the buildup of atheromatous plaque. At onset there are usually no symptoms, but if they develop, symptoms generally begin around middle age.\\
What type of animal is a jellyfish? | Jellyfish - Simple English Wikipedia, the free encyclopedia: Jellyfish are animals of the phylum Cnidaria. They are a monophyletic clade, the Medusozoa. Most of them live in the oceans, in salt water, where they eat small sea animals like plankton and little fish, and float in the sea.\\
Is an invertebrate susceptible to atherosclerosis? | Atherosclerosis: Atherosclerosis is a pattern of the disease arteriosclerosis in which the wall of the artery develops abnormalities, called lesions.\\
What is atherosclerosis? | Atherosclerosis: Atherosclerosis is a pattern of the disease arteriosclerosis in which the wall of the artery develops abnormalities, called lesions. These lesions may lead to narrowing due to the buildup of atheromatous plaque. At onset there are usually no symptoms, but if they develop, symptoms generally begin around middle age.\\
Do jellyfish have arteries? | Jellyfish: Jellyfish are mainly free-swimming marine animals with umbrella-shaped bells and trailing tentacles, although a few are anchored to the seabed by stalks rather\\
What is atherosclerosis? | Atherosclerosis: Atherosclerosis is a pattern of the disease arteriosclerosis in which the wall of the artery develops abnormalities, called lesions. These lesions may lead to narrowing due to the buildup of atheromatous plaque. At onset there are usually no symptoms, but if they develop, symptoms generally begin around middle age.\\
Do jellyfish have arteries? | Jellyfish: Jellyfish are mainly free-swimming marine animals with umbrella-shaped bells and trailing tentacles, although a few are anchored to the seabed by stalks rather\\
So jellyfish don't have atherosclerosis because they don't have arteries? | Jellyfish: A free-swimming marine coelenterate that is the sexually reproducing form of a hydrozoan or scyphozoan and has a nearly transparent saucer-shaped body and\\\\
Question:\\\\
Is a jellyfish safe from atherosclerosis?\\
Answer:\\
Jellyfish do not have a circulatory system and therefore do not have arteries. Atherosclerosis is a disease in which plaque builds up on the walls of the arteries.\\
So the answer is: Yes.\\
...\\
\#
\\
Context:\\
What is arithmetic? | Arithmetic: Arithmetic is an elementary part of mathematics that consists of the study of the properties of the traditional operations on numbers—addition, subtraction, multiplication, division, exponentiation, and extraction of roots.\\
What is the evidence that Neanderthals used arithmetic? | Neanderthal: In 2012, British-American geneticist Graham Coop hypothesised that they instead found evidence of a different archaic human species interbreeding with modern\\
Is there any reason to believe that Neanderthals could not have used arithmetic? | Neanderthal: A large part of the controversy stems from the vagueness of the term "species", as it is generally used to distinguish two genetically isolated populations, but\\
What is arithmetic? | Arithmetic: Arithmetic is an elementary part of mathematics that consists of the study of the properties of the traditional operations on numbers—addition, subtraction, multiplication, division, exponentiation, and extraction of roots.\\
Do we have any evidence that Neanderthals could manipulate numbers? | Neanderthal: Neanderthals also written as Neandertals, are an extinct species or subspecies of archaic humans who lived in Eurasia until about 40,000 years ago.\\
What is arithmetic? | Neanderthal: Neanderthals also written as Neandertals, are an extinct species or subspecies of archaic humans who lived in Eurasia until about 40,000 years ago.\\
What is the earliest evidence of arithmetic? | Mathematics: It is in Babylonian mathematics that elementary arithmetic (addition, subtraction, multiplication, and division) first appear in the archaeological record. The Babylonians also possessed a place-value system and used a sexagesimal numeral system which is still in use today for measuring angles and time.\\
Did Neanderthals live before the ancient Babylonians? | Neanderthal: Neanderthals also written as Neandertals, are an extinct species or subspecies of archaic humans who lived in Eurasia until about 40,000 years ago.  Pre- and early Neanderthals, living before the Eemian interglacial\\\\
Question:\\
Did Neanderthals use arithmetic?\\\\
Answer:\\
There is no evidence that Neanderthals used arithmetic. The earliest evidence of arithmetic dates back to the ancient Babylonians who lived after the Neanderthals.\\
So the answer is: No.\\
\#
\\\bottomrule
\end{tabular}
\caption{Instruction and exemplars for the \strategyqa{} prompt for \mtec{} and \stec{} reasoning over retrieved evidence.}
\label{fig:strategyqa_entailment_c_prompt}
\end{figure*}

\end{document}